\documentclass{tlp}
\usepackage{aopmath}
\usepackage{latexsym}
\usepackage{times}
\usepackage{amsfonts}
\usepackage{graphicx}
\usepackage{amssymb}
\usepackage{enumerate}
\usepackage{algorithmic}
\usepackage{algorithm}
\usepackage{hyperref}
\usepackage{color}

\newcommand{\Return}{\\\bf{return}}

\newcommand{\rto}{\rightarrow}
\newcommand{\lto}{\leftarrow}
\newcommand{\lrto}{\leftrightarrow}
\newcommand{\Rto}{\Rightarrow}
\newcommand{\Lto}{\Leftarrow}
\newcommand{\LRto}{\Leftrightarrow}

\newcommand{\Not}{not \,}

\newcommand{\Pos}{\textit{Pos}}

\newcommand{\Neg}{\textit{Neg}}

\newcommand{\comp}{C\!O\!M\!P}

\newcommand{\lfp}{\textit{lfp}}

\newcommand{\HB}{\textit{HB}}
\newcommand{\wLF}{\textit{wLF}}
\newcommand{\sLF}{\textit{sLF}}

\newcommand{\cLF}{\textit{cLF}}

\newtheorem{example}{Example} 

\begin{document}
\bibliographystyle{acmtrans}

\long\def\comment#1{}

\title{Loop Formulas for Description Logic Programs\footnote{This is the full version of \cite{Yisong:ICLP:2010}.}}

\author[Y. Wang et al.]
{YISONG WANG\\
Department of Computer Science, Guizhou University, Guiyang, China\\
Department of Computing Science,University of Alberta, Canada
 \and JIA-HUAI YOU, LI YAN YUAN\\
 Department of Computing Science,University of Alberta, Canada
 \and YI-DONG SHEN\\
 State Key Laboratory of Computer Science Institute of
Software, Chinese Academy of Sciences, China}
%

\pagerange{\pageref{firstpage}--\pageref{lastpage}}
\submitted{6 February 2010}
\revised{7 April 2010}
\accepted{14 May 2010}
\maketitle

\label{firstpage}

\begin{abstract}
Description Logic Programs (dl-programs) proposed by Eiter et al.
constitute an elegant yet powerful formalism for the integration of answer set programming with description logics, for the
Semantic Web. In this paper, we generalize the notions of completion
and loop formulas of logic programs to description logic programs
and show that the answer sets of a dl-program can be precisely
captured by the models of its completion and loop formulas.
Furthermore, we propose a new, alternative semantics
for dl-programs, called the {\em canonical answer set semantics}, which is
defined by the models of completion that satisfy what are called
canonical loop formulas. A
desirable property of canonical answer sets is that they are free of
circular justifications. Some properties of canonical answer sets
are also explored.
\end{abstract}

\section{Introduction}
Logic programming under the answer set semantics (ASP) is
a nonmonotonic reasoning
paradigm for declarative problem
solving \cite{marek99,Niemela99}.
Recently,
there have been extensive interests in combining ASP with other
computational and reasoning paradigms.
\comment{(e.g.,
\cite{DBLP:conf/ijcai/MotikR07,DBLP:journals/ai/EiterILST08,Mellarkod:AMAI2008}).}
One of the main interests in this direction is the integration of ASP
with ontology reasoning, for the Semantic Web.

The Semantic Web
is an evolving development of the World Wide
Web in which the meaning of information and services on the web are
defined, so that the web content can be precisely understood and
used by agents \cite{Berners:Science:2001}. For this purpose, a
layered structure including the Rules Layer built on top of the
Ontology Layer has been  recognized as a fundamental framework.
Description Logics (DLs) \cite{Badder:Handbook:DL:2007} provide a
formal basis for the Web Ontology Language which is the standard of
the Ontology Layer \cite{OWL2-overview}.

Adding nonmonotonic rules to the Rules Layer would allow default reasoning with ontologies.
For example, we know that most {\em natural kinds} do not have a
clear cut definition. For instance, a precise definition of {\em
scientist} seems to be difficult by enumerating what a scientist is,
and does. Though we can say that a scientist possesses expert
knowledge on the subject of his or her investigation, we still
need a definition of expert knowledge,
which cannot be defined quantitatively. Using nonmonotonic
rules, we can perform default, typicality
reasoning over categories, concepts,
and roles. The integration of DLs and (nonmonotonic) rules has been
extensively investigated as a crucial problem in the study of the
Semantic Web, such as {\em Semantic Web Rule Language} (SWRL)
\cite{DBLP:Horrocks:WWW:2004}, {\em MKNF knowledge base}
\cite{Motik:JACM:2010}, and {\em Description Logic Programs}
({\em dl-programs}) \cite{DBLP:journals/ai/EiterILST08}.

There are different approaches to the integration of ASP with
description logics.
The focus of this paper is on the
approach based on dl-programs. Informally, a dl-program is a pair
$(O,P)$, where $O$ is a DL knowledge base and $P$ is a logic program whose rule bodies may contain
queries, embedded in {\em dl-atoms}, to the knowledge base $O$. The
answer to such a query depends on inferences by rules over the DL
knowledge base $O$. In this way, rules are built on top of
ontologies. On the other hand, ontology reasoning is also enhanced,
since it depends not only on $O$ but also on inferences using
(nonmonotonic) rules.
Two semantics for dl-programs have been proposed, one of which is based on
{\em strong answer sets} and the other based on {\em weak answer
sets}.

In this paper, we generalize the notions of completion and loop
formulas of logic programs \cite{LinZhao:assat}
to dl-programs and show that weak and strong answer sets
of a dl-program can be captured precisely by the models of its
completion and the corresponding loop formulas.
This provides not only a semantic
characterization of answer sets for dl-programs but also an
alternative mechanism for answer set computation, using a dl-reasoner and a
SAT solver.

As commented by \cite{DBLP:journals/ai/EiterILST08}, the reason to introduce
strong answer sets is because some weak answer sets seem counterintuitive due
to ``self-supporting'' loops. Recently however,
one of the co-authors of this paper,
Yi-Dong Shen, discovered that
strong answer sets may also possess self-supporting loops,
and a detailed analysis leads to the conclusion that
the problem cannot be easily fixed
by an alternative definition of {\em reduct}, since
the reduct of dl-atoms may not be able to
capture dynamically generated self-supports arising from the integrated
context.

The solution proposed in this paper is to use loop formulas as a way
to define answer sets for dl-programs that are free of
self-supports. Thus, we define what are called {\em canonical loops}
and {\em canonical loop formulas}. Given a dl-program, the models of
its completion satisfying the canonical loop formulas
constitute a new class of answer sets, called  {\em canonical answer
sets}, that are minimal and noncircular.

The paper is organized as follows. In the next section, we
recall the basic definitions of description
logics and dl-programs. In
Section~\ref{Completion-and-Loop-Formulas}, we define
completion, weak and strong loop formulas for dl-programs. The new
semantics of dl-programs based on canonical loop formulas is given
in Section~\ref{Canonical-Answer-Sets}. Section~\ref{related-work}
discusses related work, and finally
Section~\ref{Discussion-and-Future-Work} gives
concluding remarks.

\section{Preliminaries}
\label{preliminaries}
In this section, we briefly review the basic notations for
description logics and description logic programs
\cite{DBLP:journals/ai/EiterILST08}.

\vspace{-.05in}
\subsection{Description logics}

In principle,
the description logics
employed in description logic programs can be arbitrary, with
the restriction that the underlying entailment relation is decidable.
Due to space limitation, we introduce
the basic description logic ${\cal ALC}$ \cite{Badder:Handbook:DL:2007},
instead of the description logics ${\cal SHIF}$ and ${\cal SHOIN}$
described in \cite{DBLP:journals/ai/EiterILST08}. The notations introduced here will be used throughout the paper, particularly the entailment relation
$O\models F$, given at the end of this subsection.

For the language ${\cal ALC}$, we assume a
vocabulary $\Psi=({\bf A\cup R,I})$, where ${\bf A, R}$ and ${\bf I}$ are pairwise disjoint (denumerable)
sets of {\em atomic concepts}, {\em roles} (including
equality $\approx$
and inequality $\not\approx$), and {\em individuals} respectively.
The {\em concepts} of ${\cal ALC}$ are defined as follows:
$$C,D\longrightarrow A |\top | \bot | \neg C | C \sqcap D | C \sqcup D | \forall R.C | \exists R.C$$
where $A$ is an atomic concept and $R$ is a role. The {\em assertions} of
${\cal ALC}$ are of the forms $C(a)$ or $R(b,c)$, where $C$ is a concept, $R$ is
a role, and $a,b,c$ are individuals. An {\em inclusion axiom} of ${\cal ALC}$ has the form
$C\sqsubseteq D$ where $C$ and $D$ are concepts. A {\em description knowledge base} (or {\em ontology})
of ${\cal ALC}$ is a set of inclusion axioms and assertions of ${\cal ALC}$.

The semantics of ${\cal ALC}$ is defined by  translating
to first-order logic and then using classical first-order interpretations as its semantics.
Informally, let the transformation be $\tau$: (1) $\tau(A)=A(x)$, $\tau(R)=R(x,y)$ where
$A$ is an atomic concept and $R$ a role; (2) $\tau(\forall R.C)=\forall x. R(y,x)\supset\tau(C)(x)$,
and $\tau(\exists R.C)=\exists x.R(y,x)\wedge\tau(C)(x)$; (3) $\tau(\neg C)=\neg \tau(C)(x)$,
$\tau(C\sqcap D)=\tau(C)(x)\wedge \tau(D)(x)$, and $\tau(C\sqcup D)=\tau(C)(x)\vee\tau(D)(x)$;
(4) $\tau(A(a))=A(a)$, $\tau(R(b,c))=R(b,c)$;
(5) $\tau(C\sqsubseteq D)=\forall x. \tau(C)(x)\supset\tau(D)(x)$.
Then, the
semantics of ${\cal ALC}$ follows from that of first-order logic,
so is the entailment relation $O\models F$, for a description
knowledge base $O$ and an assertion or inclusive axiom $F$.

\vspace{-.05in}
\subsection{Description logic programs}
Let $\Phi=(\mathcal {P,C})$ be a first-order vocabulary with nonempty finite sets $\cal C$
and $\cal P$ of constant symbols and predicate symbols respectively such that
$\cal P$ is disjoint from ${\bf A\cup R}$ and $\cal C\subseteq \bf I$.
{\em Atoms} are formed from the symbols in $\cal P$ and $\cal C$ as usual.

A {\em dl-atom} is an expression of the form
\begin{equation}\label{dl:atom}
  DL[S_1\ op_1\ p_1, \ldots,S_m\ op_m\ p_m;Q](\vec t),\ \  (m\ge 0)
\end{equation}
where
\begin{itemize}
  \item each $S_i$ is either a concept, a role\comment{ or its negation\footnote{
  We allow the negation of a role for convenience, so that
  we can replace ``$S\odot p$" with an equivalent form ``$\neg S\oplus p$" in dl-atoms. The negation of a role is not explicitly present in
\cite{DBLP:journals/ai/EiterILST08}. },} or a special
  symbol in $\{\thickapprox,\not\thickapprox\}$;
  \item $op_i\in\{\oplus,\odot,\ominus\}$;
  \item $p_i$ is a unary predicate symbol in $\cal P$ if $S_i$ is a concept, and a binary predicate symbol in $\cal P$
   otherwise. The $p_i$s are called {\em input predicate symbols};
  \item $Q(\vec t)$ is a {\em dl-query}, i.e., either (1) $C(t)$ where $\vec t=t$;
    (2) $C\sqsubseteq D$ where $\vec t$ is an empty  argument list;
    (3) $R(t_1,t_2)$ where $\vec t=(t_1,t_2)$;
    (4) $t_1\thickapprox t_2$ where $\vec t=(t_1,t_2)$;
    or their negations, where $C$ and $D$ are concepts, $R$ is a role, and $\vec t$
  is a tuple  of constants.
\end{itemize}

The precise meanings of $\{\oplus, \odot, \ominus\}$ will be defined shortly.
Intuitively, $S \oplus p$ (resp. $S \odot p$)
extends $S$ (resp. $\neg S$)
by the extension of $p$, and
$S \ominus p$
constrains $S$ to $p$.

For example, suppose the interface is such that if
any individual $x$ is registered for a course (the information from outside an ontology)
then $x$ is a student ($x$ may not be a student by the ontology
before this communication), and we
query if $a$ is a student. We can then write
the dl-atom $DL[Student \oplus registered; Student](a)$.
Similarly, $DL[Student \ominus registered;
\neg Student \sqcap \neg Employed](a)$
queries if $a$ is not a student nor employed,
with the
ontology enhancement that if we cannot show $x$ is registered,
then $x$ is not a student.

A {\em dl-rule} (or simply a {\em rule}) is an expression of the form
\begin{equation}\label{normal:rule}
  A\lto B_1,\ldots,B_m,\Not B_{m+1},\ldots,\Not B_n, (n\ge m\ge 0)
\end{equation}
where $A$ is an atom, each $B_i ~(1\leq i\leq n)$ is an atom\footnote{Different from that
of \cite{DBLP:journals/ai/EiterILST08}, we consider ground atoms
instead of literals for convenience.} or a dl-atom. We refer to $A$ as its {\em
head}, while the conjunction of $B_i(1\leq i\leq m)$ and $\Not
B_j~(m+1\leq j\leq n)$ is its {\em body}. For convenience, we may
abbreviate a rule in the form (\ref{normal:rule}) as
\begin{equation}\label{normal:rule:set}
  A\lto \Pos,\Not \Neg
\end{equation}
where $\Pos=\{B_1,\ldots,B_m\}$ and $\Neg=\{B_{m+1},\ldots,B_n\}$.
Let $r$ be a rule of the form (\ref{normal:rule:set}).
If $\Neg=\emptyset$
and $\Pos=\emptyset$, $r$ is a fact and we may write it as ``$A$" instead of
``$A\lto$".
A {\em description logic program} ({\em dl-program})
$\mathcal K=(O,P)$ consists of a DL knowledge base $O$
and a finite set $P$ of dl-rules. In what follows we assume the vocabulary
of $P$ is implicitly given by the constant symbols and predicates symbols occurring in $P$,
unless stated otherwise.

Given a dl-program $\mathcal K=(O,P)$, the {\em Herbrand base} of
$P$, denoted by $\textit{HB}_P$, is the set of atoms formed from the
predicate symbols in $\mathcal P$ occurring in $P$ and the constant
symbols in $\mathcal C$ occurring in $P$. An {\em interpretation}
$I$ (relative to $P$) is a subset of $\HB_P$. Such an $I$ is a {\em
model} of an atom or dl-atom $A$ under $O$, written $I\models_OA$,
if the following holds:
\begin{itemize}
  \item if $A\in\HB_P$, then $I\models_OA$ iff $A\in I$;
  \item if $A$ is a dl-atom $DL(\lambda;Q)(\vec t)$ of the form (\ref{dl:atom}), then $I\models_OA$ iff $
    O(I;\lambda)\models Q(\vec t)$ where
    $O(I;\lambda)=O\cup\bigcup_{i=1}^{m}A_i(I)$\mbox{ and, for $1\leq i\leq m$,}
    \[A_i(I)=\left\{
    \begin{array}{ll}
        \{S_i(\vec e)|p_i(\vec e)\in I\}, & \hbox{if $op_i=\oplus$;} \\
        \{\neg S_i(\vec e)|p_i(\vec e)\in I\}, & \hbox{if $op_i=\odot$;} \\
        \{\neg S_i(\vec e)|p_i(\vec e)\notin I\}, & \hbox{if $op_i=\ominus$;}
    \end{array}
    \right.
    \]
\end{itemize}
where $\vec e$ is a tuple of constants over $\mathcal C$.
The interpretation $I$ is a {\em model} of a dl-rule of the form
(\ref{normal:rule:set}) iff $I\models_O B$ for any $B\in\Pos$ and
$I\not\models_OB'$ for any $B'\in\Neg$ implies $I\models_OA$.
$I$ is a {\em model} of a dl-program $\mathcal
K=(O,P)$, written $I\models_O\mathcal K$, iff $I$ is a model of each
rule of $P$. $I$ is a {\em supported model} of $\mathcal K=(O,P)$
iff, for any $h\in I$, there is a rule ($h\lto \Pos,\Not\Neg$) in
$P$ such that $I\models_O A$ for any $A\in\Pos$ and $I\not\models_O
B$ for any $B\in\Neg$.

A dl-atom $A$ is {\em monotonic} relative to a dl-program $\mathcal
K=(O,P)$ if $I\models_ OA$ implies $I'\models_ OA$, for all
$I\subseteq I'\subseteq \textit{HB}_P$, otherwise $A$ is {\em
nonmonotonic}. It is clear that if a dl-atom does not mention
$\ominus$ then it is monotonic. However, a dl-atom may be monotonic
even if it mentions $\ominus$. E.g., the dl-atom $DL[S\odot
p,S\ominus p;\neg S](a)$ is monotonic (which is a tautology).
Clearly, the $\ominus$ operator is the only one that may cause a
dl-atom to be nonmonotonic. Thus one has no reason to use $\ominus$
in monotonic dl-atoms. It is a reasonable assumption that we can
rewrite a monotonic dl-atom into an equivalent one without using
$\ominus$ at all.

We use $DL_P$ to denote the set of all dl-atoms that occur in $P$,
$DL_P^{+}\subseteq DL_P$ to denote the set of monotonic dl-atoms, and
$DL_P^{?}=DL_P\setminus DL_P^{+}$. A dl-program $\mathcal K=(O,P)$ is
{\em positive} if (i) $P$ is ``not"-free, and (ii) every dl-atom is
monotonic relative to $\mathcal K$. It is evident that if a
dl-program $\mathcal K$ is positive, then $\mathcal K$ has a
(set inclusion)
least
model.

\subsection{Strong and weak answer sets}
Let $\mathcal K=(O,P)$ be a dl-program. The {\em strong
dl-transform} of $\mathcal K$ relative to $O$ and an interpretation
$I\subseteq \textit{HB}_P$, denoted by $\mathcal K^{s,I}$, is the
positive dl-program $(O,sP^I_O$), where $sP^I_O$ is obtained from
$P$ by deleting:
\begin{itemize}
  \item the dl-rule $r$ of the form (\ref{normal:rule}) such that either
    $I\not\models_OB_i$ for some $1\leq i\leq m$ and
    $B_i\in DL_P^?$, or $I\models_OB_j$ for some $m+1\leq j\leq n$; and
  \item the nonmonotonic dl-atoms and $\Not A$ from the remaining dl-rules where $A$ is an
  atom or dl-atom.
\end{itemize}
The interpretation $I$ is a {\em strong answer set} of $\mathcal K$
if it is the least model of $\mathcal K^{s,I}$.

The {\em weak
dl-transform} of $\mathcal K$ relative to $O$ and an interpretation
$I\subseteq HB_P$, denoted by $\mathcal K^{w,I}$, is the positive
dl-program $(O,wP^I_O)$, where $wP_O^I$ is obtained from $P$ by
deleting:
\begin{itemize}
  \item the dl-rules of the form (\ref{normal:rule}) such that either
    $I\not\models_OB_i$ for some $1\leq i\leq m$ and
    $B_i\in DL_P$, or $I\models_OB_j$ for some $m+1\leq j\leq n$; and
  \item the dl-atoms and $\Not A$ from the remaining dl-rules where
  $A$ is an atom or dl-atom.
\end{itemize}
The interpretation $I$ is a {\em weak answer set} of $\mathcal K$ if
$I$ is the least model of $\mathcal K^{w,I}$.

\begin{example}\label{exam:dl:program:1}
Consider the following dl-programs:
\begin{itemize}
\item
  $\mathcal K_0=(O,P_0)$ where $O=\{c\sqsubseteq c'\}$ and $P_0=\{w(a)  \!\lto \! DL[c\oplus p;c'](a); p(a) \leftarrow\}$. For this dl-program to make
some sense, let's image this situation: $c'$ and $c$ are classes of
  good conference papers and
  ICLP papers respectively,
$p(x)$ means that $x$ is a paper in the TPLP special issue of ICLP 2010,
$w(x)$ means that $x$ is worth reading, and
$a$ stands for ``this paper''. Note that $c$ and $c'$ are concepts in
$O$, and $p$ and $w$ are predicates outside of $O$.
The communication is through the dl-rule,
$w(a)\lto DL[c\oplus p;c'](a)$, which says that
if ``this paper'' is a
good conference paper,
given that
any paper in
the TPLP special issue of ICLP 2010 is an ICLP paper and ICLP papers
are good conference papers (by the knowledge in $O$),
then it is worth reading.
$\mathcal K_0$
has exactly one
strong answer set
$\{p(a), w(a)\}$,
which
is also its unique
weak answer set.

\item
Now, suppose someone writes
$\mathcal K_1=(O,P_1)$ where $O=\{c\sqsubseteq c'\}$ and
$P_1=\{p(a)\lto DL[c\oplus p;c'](a)\}$.
This program has a
  unique strong answer set $I_1 = \emptyset$ and two weak answer sets $I_1$ and $I_2 = \{p(a)\}$.
It can be seen that there is a circular justification
in the weak answer set $I_2$:
that ``this paper'' is in the TPLP special issue of ICLP 2010
is justified by its being in it.

The interested reader may verify the following.
By the definition of $\oplus$,
$O(I_2;c\oplus p)=O\cup\{c(a)\}$, and clearly $O\not\models c'(a)$ and
  $\{c(a),c\sqsubseteq c'\}\models c'(a)$. So the weak dl-transform
relative to $O$ and $I_2$
is
${\cal K}_1^{w, I_2}=(O,\{p(a) \leftarrow \})$. Since
$I_2$ coincides with the least model of $\{p(a) \leftarrow \}$, it is
a weak answer set of
$\mathcal K_1$.
Similarly, one can verify that the strong dl-transform relative to $O$ and
$I_2$ is
${\cal K}_1^{s,I_2}=(O,P_1)$. Its least model is the empty set,
so $I_2$ is not a strong answer set of $\mathcal K_1$.

\item
  $\mathcal K_2=(O,P_2)$ where $O=\emptyset$ and $P_2=\{p(a)\lto DL[c\oplus p, b\ominus q;c\sqcap\neg b](a)\}$.
  Both $\emptyset$ and $\{p(a)\}$ are strong and weak answer sets of the
  dl-program.

\item
  $\mathcal K_3=(\emptyset,P_3)$ where $P_3=\{p(a)\lto DL[c\odot p, b\ominus q;\neg c\sqcap\neg b](a)\}$.
  $\emptyset$ and $\{p(a)\}$ are both strong and weak answer sets of the dl-program.
  \item $\mathcal K_4=(\emptyset,P_4)$ where $P_4=\{p(a)\lto DL[c\ominus p; \neg c](a)\}$.
  $\mathcal K_4$ has no weak answer set, and thus it has no strong answer set either.
\end{itemize}
\end{example}

These dl-programs show that strong (and weak) answer sets may not be
(set inclusion) minimal. It has been shown that if a
dl-program contains no nonmonotonic dl-atoms then its strong answer
sets are minimal \cite{DBLP:journals/ai/EiterILST08}. However, this
does not hold for weak answer sets as shown by the dl-program
$\mathcal K_1$ above, even if it is positive. It is known
that strong answer sets are always weak answer sets, but
not vice
versa \cite{DBLP:journals/ai/EiterILST08}.
\vspace{-.1cm}
\section{Completion and Loop Formulas}
\label{Completion-and-Loop-Formulas}
\label{loop} In this section, we define completion, characterize
weak and strong answer sets by
loop formulas, and outline an alternative method of computing
weak and strong answer sets.
\vspace{-.1cm}
\subsection{Completion}
Given a dl-program $\mathcal K=(O,P)$, we assume an underlying
propositional language $\mathcal L_\mathcal K$, such that the
propositional atoms of $\mathcal L_\mathcal K$ include the atoms and
dl-atoms occurring in $P$. The {\em formulas} of $\mathcal L_\mathcal K$
are defined as usual using the connectives $\neg,\wedge,\vee, \supset$
and $\lrto$. The {\em dl-interpretations} (or simply {\em interpretations} if
it is clear from context) of the language $\mathcal
L_\mathcal K$ are the interpretations relative to $P$, i.e., the
subsets of $\HB_P$. For a formula $\psi$ of $\mathcal L_\mathcal K$
and an interpretation $I$ of $\mathcal L_\mathcal K$, we say $I$ is
a {\em model} of $\psi$ relative to $O$, denoted $I\models_O\psi$,
whenever (i) if $\psi$ is an atom, then $\psi\in I$;
  (ii) if $\psi$ is a dl-atom, then $I\models_O\psi$; and
  (iii) the above is extended in the usual way to arbitrary formulas of
$\mathcal L_\mathcal K$.

Let $\mathcal K=(O,P)$ be a dl-program and $h$ an atom in $\HB_P$.
The {\em completion} of $h$ (relative to $\mathcal K$), written
$\comp(h,\mathcal K)$, is the following formula of $\mathcal
L_\mathcal K$:
\begin{eqnarray*}
  h\lrto \bigvee_{1\leq i\leq n} \left(\bigwedge_{A\in \Pos_i} A\wedge\bigwedge_{B\in\Neg_i}\neg B\right),
\end{eqnarray*}
where $(h\lto \Pos_1,\Not\Neg_1),\ldots,(h\lto\Pos_n,\Not\Neg_n)$
are all the rules in $P$ whose heads are the atom $h$. The {\em
completion} of $\mathcal K$, written $\comp(\mathcal K)$, is the
collection of completions of all atoms in $\HB_P$.

Recall that a model $M \subseteq \HB_P$
of a
dl-program $\mathcal K=(O,P)$ is a {\em supported model} if
for any atom $a \in M$, there is a rule in $P$ whose head is $a$ and whose body
is satisfied by $M$.

\begin{proposition}\label{prop:completion:supported}
  Let $\mathcal K=(O,P)$ be a dl-program and $I$ an interpretation of $P$.
  Then $I$ is a supported model of $\mathcal K$
  if and only if $I\models_O\comp(\mathcal K)$.
\end{proposition}

\begin{proposition}\label{prop:AS:supported:model}
Every weak (resp. strong) answer
  set of a dl-program $\mathcal K$
is a supported model of $\mathcal K$.
\end{proposition}

\subsection{Weak loop formulas}
\label{weak-loop}
In order to capture weak answer sets of dl-programs using completion
and loop formulas, we define weak loops. Formally, let
$\mathcal K=(O,P)$ be a dl-program. The {\em weak positive
dependency graph} of $\mathcal K$, written $G_\mathcal K^w$, is the
directed graph $(V,E)$, where $V=\HB_P$ (note that a dl-atom is not in $V$), and $(u,v)\in E$ if there is a dl-rule of the form (\ref{normal:rule}) in
$P$ such    that $A=u$ and $B_i=v$ for  some
$i~(1\leq i\leq m)$. A nonempty subset $L$ of $\HB_P$ is a {\em weak
loop} of $\mathcal K$ if there is a cycle in $G^w_\mathcal K$ which
goes through only and all the nodes in $L$.

Given a weak loop $L$ of a dl-program $\mathcal K=(O,P)$, the {\em
weak loop formula} of $L$ (relative to $\mathcal K$), written
$\wLF(L,\mathcal K)$, is the following formula of $\mathcal
L_\mathcal K$:
\begin{eqnarray*}
  \bigvee L\supset \bigvee_{1\leq i\leq n}\left(\bigwedge_{A\in\Pos_i}A\wedge\bigwedge_{B\in\Neg_i}\neg B\right)
\end{eqnarray*}
where $(h_1\lto\Pos_1,\Not\Neg_1),\ldots,
(h_n\lto\Pos_n,\Not\Neg_n)$ are all the rules in $P$ such that
$h_i\in L$ and $\Pos_i\cap L=\emptyset$ for any $i~(1\leq i\leq n)$.

\begin{theorem}
  Let $\mathcal K=(O,P)$ be a dl-program and $I$ an interpretation of $P$. Then $I$ is a weak answer set of $\mathcal K$
  if and only if $I\models_O\comp(\mathcal K)\cup \textit{wLF}(\mathcal K)$, where $\textit{wLF}(\mathcal K)$
  is the set of weak loop formulas of all weak loops of  $\mathcal K$.
\label{weak}
\end{theorem}

\subsection{Strong loop formulas}
Let $\mathcal K=(O,P)$ be a dl-program. The {\em strong positive
dependency graph} of $\mathcal K$, denoted by $G^s_\mathcal K$, is
the directed graph $(V,E)$, where $V=\textit{HB}_P$ and $(p(\vec
c),q(\vec c'))\in E$ if there is a rule of the form
(\ref{normal:rule}) in $P$ such that, (1) $A=p(\vec c)$ and, (2) for
some $i~(1\leq i\leq m)$, either
\begin{itemize}
  \item $B_i=q(\vec c')$, or
  \item $B_i$ is a monotonic dl-atom mentioning the predicate $q$ and $\vec c'$ is
  a tuple of constants matching the arity of $q$.
(If this condition is ignored then
  it becomes the definition of weak positive dependency graph.)
\end{itemize}
A nonempty subset $L$ of $\textit{HB}_P$ is a {\em strong loop} of
$\mathcal K$ if there is a cycle in $G^s_\mathcal K$ which passes
only
and all the nodes in $L$.

To define strong loop formulas of a dl-program $\mathcal K=(O,P)$,
we need to extend the vocabulary $\Phi$, such that, for any
predicate symbol $p$ and a nonempty set of atoms $L$, $\Phi$
contains the predicate symbol $p_L$ that has the same arity as that
of $p$.

Let $L$ be a nonempty set of atoms, $A=DL[\lambda;Q](\vec t)$ be a
dl-atom. The {\em irrelevant formula} of $A$ relative to $L$,
written by $\textit{IF}(A,L)$, is the conjunction of (1)
$DL[\lambda_L;Q](\vec t)$, where $\lambda_L$ is obtained from
$\lambda$ by replacing each predicate symbol $p$ with $p_L$ whenever
$p$ appears in both $\lambda$ and $L$ and, (2) for each predicate
symbol $p$ mentioned in both $\lambda$ and $L$, the
instantiation on $\mathcal C$ \cite{CLWZ06} of the formula:
\begin{eqnarray}\label{IF:formula}
  \forall \vec X.\left[ p_L(\vec X)\lrto \left(p(\vec X)\wedge\bigwedge_{p(\vec c)\in L}\vec X\neq \vec
  c\right)\right]
\end{eqnarray}
where $\vec X$ is a tuple of distinct variables matching the arity
of $p$, and $\vec X\neq \vec c$ stands for $\neg(\vec X=\vec c)$,
i.e., $\neg (x_1=c_1\wedge\ldots\wedge x_k=c_k)$ if $\vec
X=(X_1,\ldots,X_k)$ and $\vec c=(c_1,\ldots, c_k)$. Please note that,
the instantiation of a formula $\forall x\cdot\psi$ on a finite set $D$ of constants is the
formula $\bigwedge_{d\in D}\psi[x/d]$, in which $c=c$ (resp., $c=c'$) is replaced
with $\top$ (\textit{true}) (resp., $\bot$ (\textit{false})), where
$c$ and $c'$ are two distinct constants. In what follows, we identify the formula
(\ref{IF:formula}) with its instantiation whenever it is clear from its context, unless otherwise stated.

 For instance,
let $A=DL[c\oplus p;c](a)$ and $L=\{p(a),p(b)\}$. Then
$IF(A,L)$ is the formula:
\begin{eqnarray*}
  DL[c\oplus p_L;c](a)\wedge (p_L(a)\lrto p(a)\wedge a\neq a)\wedge(p_L(b)\lrto p(b)\wedge a\neq b)
\end{eqnarray*}
which is equivalent to
\[ DL[c\oplus p_L;c](a)\wedge \neg p_L(a)\wedge (p_L(b)\lrto p(b)).\]
Intuitively, the irrelevant formula of $A$ relative to $L$ says that
the truth of $A$ only depends on the truth of the atoms not in $L$.

We are now in a position to define strong loop formulas. Let $L$ be
a strong loop of $\mathcal K=(O,P)$. The {\em strong loop formula}
of $L$ (relative to $\mathcal K$), written $\sLF(L,\mathcal K)$, is
the following formula of $\mathcal{L_K}$:
\begin{eqnarray*}
  \bigvee L\supset \bigvee_{1\leq i\leq n}\left(\bigwedge_{A\in\Pos_i}\gamma(A,L)\wedge\bigwedge_{B\in\Neg_i}\neg B\right)
\end{eqnarray*}
where
\begin{itemize}
  \item $(h_1\lto \Pos_1,\Not\Neg_1),\ldots, (h_n\lto\Pos_n,\Not\Neg_n)$ are all the rules in $P$
    such that $h_i\in L$ and  $\Pos_i\cap L=\emptyset$ for all $i~(1\leq i\leq n)$,
  \item $\gamma(A,L)=\textit{IF}(A,L)$ if $A$ is a monotonic dl-atom, and $A$ otherwise.
\end{itemize}

In general, we
have to recognize the monotonicity of dl-atoms in order to construct strong loops of
dl-programs. In this sense, the strong loops and strong loop formulas are defined semantically.
If a dl-atom does not mention the operator $\ominus$ then it is obviously monotonic. Thus for the
class of dl-programs in which no monotonic dl-atoms mention $\ominus$, the strong loops and strong loop formulas are
given syntactically, since it is sufficient to determine the monotonicity of
a dl-atom by checking whether it contains the operator $\ominus$.

\begin{example}\label{exam:3}
  Let $\mathcal K=(\emptyset,P)$ be a dl-program where $P$ consists of
$$
p(a)  \lto DL[c\oplus p;c](a); ~~~~~~~~
p(a)   \lto \Not DL[c\oplus p;c](a).
$$
The dl-program $\mathcal K$ has a unique
  strong loop $L=\{p(a)\}$, but doesn't have any weak loops. Its completion is the formula:
  \[p(a)\lrto DL[c\oplus p;c](a)\vee \neg DL[c\oplus p; c](a)\]
  which equals to the formula $p(a)\lrto \top$, i.e., $p(a)$. Note that, the strong
  loop formula $\sLF(L,\mathcal K)$ is the formula:
  \begin{eqnarray*}
   p(a)\supset \left[\begin{array}{l}
    DL[c\oplus p_L;c](a)\wedge (p_L(a)\lrto p(a)\wedge a\neq a)\\
    \vee \neg DL[c\oplus p;c](a)
    \end{array}\right].
  \end{eqnarray*}
  It is clear that the interpretation $I=\{p(a)\}$ is a model of $\comp(\mathcal K)$ relative to the
  DL knowledge base $O=\emptyset$. However,
  $I\not\models_O \sLF(L,\mathcal K)$.
\comment{Therefore, $I$ is not a strong answer set of $\mathcal K$.
Actually $\mathcal K$ has no strong answer set.
Since $I$ satisfies the completion and there are no weak loops, $I$
is a weak answer set, which is the only weak answer set.}
\end{example}

\begin{theorem}\label{thm:s:Loop:Comp}
  Let $\mathcal K=(O,P)$ be a dl-program and $I$ an interpretation of $P$.
  Then $I$ is a strong answer set of $\mathcal K$
  if and only if $I'\models_O\comp(\mathcal K)\cup \sLF(\mathcal K)$, where $\sLF(\mathcal K)$ is the set
  of strong loop formulas of all strong loops of $\mathcal K$ and $I'$ is the extension of
  $I$ satisfying (\ref{IF:formula}).
\end{theorem}

Since
a weak loop of a dl-program $\mathcal K$ is also
a strong loop of $\mathcal K$,  as a by-product, our loop formula characterizations
yield an alternative proof that
strong answer sets are also weak answer sets.
\begin{proposition}
  Let $\mathcal K=(O,P)$ be a dl-program, $I$ an interpretation of $P$ and $L$ a weak loop of $\mathcal K$. Then
  we have $I'\models_O\sLF(L,\mathcal K)\supset\wLF(L,\mathcal K)$,
  where $I'$ is the extension of $I$ satisfying (\ref{IF:formula}).
\end{proposition}

\subsection{An alternative method of computing weak and strong answer sets}
Theorems \ref{weak}
and \ref{thm:s:Loop:Comp} serve as the basis for
an alternative method of computing weak
and strong answer sets using
a SAT solver, along with
a dl-reasoner
${\cal R}$ with the following property:
${\cal R}$
is sound, complete, and terminating for entailment checking.
Let $\mathcal K=(O,P)$ be a dl-program and $T = COMP(\mathcal K)$.
We replace all dl-atoms in $T$ with new propositional atoms to produce
$T'$.
Let $\xi_A$ be the new atom in $T'$, for the dl-atom $A$ in $T$, and $X$ be the set
of all such new atoms in $T'$.
Below, we outline an algorithm to compute the weak answer sets
of $\mathcal K$
(here we only describe how to compute the first such an answer set).
To compute a strong answer set, replace the word weak with strong.
\begin{itemize}
\item [(i)]
Generate a model $I$ of $T$; if there is none, then there is no weak
answer set.
\item [(ii)]
Check if $I$ is a weak answer set of $\mathcal K$,
\begin{itemize}
\item [(a)] if yes, return $I$ as a weak answer set of $\mathcal K$.
\item [(b)] if no, add a
weak loop formula into $T$ that is not satisfied by $I$ relative to
$O$, and goto (i).
\end{itemize}
\end{itemize}

To generate a model of $T$, we
compute a model $M$ of $T'$ using a SAT solver,
and then use ${\cal R}$ to check the entailment:
For any dl-atom $A$ in $T$, if $M \models \xi_A$ then
$M\models_O A$ otherwise $M\not \models_O A$.
Let $M'= M/X$.
It is not difficult to verify that $M'$ is a model of $\mathcal K$.

The strong and weak answer set semantics of dl-programs have been implemented
in a
prototype
system called
SWLP\footnote{https://www.mat.unical.it/ianni/swlp/; also see \cite{DBLP:journals/ai/EiterILST08} for the details of the implementation and interesting dl-programs}, using the ASP solver DLV and a dl-reasoner.
The main difference in the method outlined here is that we use a SAT
solver to generate candidate models, which allows to take the advantages of the state-of-the-art
SAT technology.

For strong answer sets, the construction of a strong
loop formula requires checking monotonicity of dl-atoms.
However, for the class of dl-programs mentioning no
$\ominus$, this checking is not needed and the construction of a strong loop
formula is hence tractable.

\vspace{-.03in}
\section{Canonical Answer Sets}
\label{Canonical-Answer-Sets}

\subsection{Motivation: the problem of self-support}
As commented by Eiter {\em et al}.
\cite{DBLP:journals/ai/EiterILST08}, some weak answer sets
may be considered counterintuitive because of
``self-supporting" loops. For instance, consider the weak answer set
$\{p(a)\}$ of the dl-program $\mathcal K_1$ in Example
\ref{exam:dl:program:1}. The evidence of the truth of $p(a)$ is
inferred by means of a self-supporting loop: ``$p(a)\Lto DL[c\oplus p;c'](a)\Lto
p(a)$", which involves not only the dl-atom $DL[c\oplus p;c'](a)$ but
the DL knowledge base $O$. Thus the truth
of $p(a)$ depends on the truth of itself. This self-support is excluded by the strong loop formula
of the loop $L = \{p(a)\}$. \comment{Thus the weak and strong loop formulas provide a clear
distinction between the weak and strong answer sets of dl-programs.}

Let's consider the dl-program $\mathcal K_2$ in Example \ref{exam:dl:program:1} again.
Note that $\{p(a)\}$ is a strong answer set of $\mathcal K_2$.
The truth of the atom $p(a)$ depends on the truth of $[c\sqcap \neg b](a)$ which depends
on the truth of $p(a)$ and $\neg q(a)$. Thus the truth of $p(a)$ depends on the truth of itself. The
self-supporting loop is: ``$p(a)\Lto DL[c\oplus p,b\ominus q;c\sqcap \neg b](a)\Lto (p(a)\wedge\neg q(a))$". In
this sense, some strong answer sets may be considered counterintuitive as well.

The notion of ``circular justification" was formally defined by
\cite{DBLP:Liu:ICLP:2008} to characterize self-supports for lparse
programs, which was motivated by the notion of {\em unfoundedness}
for logic programs \cite{Gelderetal1991} and logic programs with
aggregates \cite{DBLP:conf/ijcai/CalimeriFLP05}. With slight
modifications, we extend the concept of circular justification to
dl-programs. Formally, let $\mathcal K=(O,P)$ be a dl-program and
$I\subseteq\HB_P$ be a supported model of $\mathcal K$. $I$ is said
to be {\em circularly justified} (or simply {\em circular}) if there
is a nonempty subset $M$ of $I$ such that
\begin{equation}\label{eq:circular}
  I\setminus M\not\models_O\bigwedge_{A\in\Pos}A\wedge\bigwedge_{B\in\Neg}\neg B
\end{equation}
for any dl-rule ($h\lto\Pos,\Not\Neg$) in $P$ with $h\in M$ and $I\models_O
\bigwedge_{A\in\Pos}A\wedge\bigwedge_{B\in\Neg}\neg B$. Otherwise, we say that $I$
is {\em noncircular}.
Intuitively speaking,  Condition (\ref{eq:circular}) means that
the atoms in $M$ have no support from outside of $M$, i.e., they have to depend on themselves.

\begin{example}\label{exam:dl-program:3}
  Let $\mathcal K=(\emptyset,P)$ where $P$ consists of
  \[p(a)\lto\Not DL[b\ominus p;\neg b](a).\]
  It is not difficult to verify that $\mathcal K$ has two weak answer sets $\emptyset$ and $\{p(a)\}$.
  They are strong answer sets of $\mathcal K$ as well. In terms of the above  definition, $\{p(a)\}$ is circular.
\end{example}

It is interesting to note that weak answer sets allow self-supporting loops
involving any dl-atoms (either monotonic or nonmonotonic),
while strong answer sets allow self-supporting loops only involving
nonmonotonic dl-atoms and their default negations.
These considerations motivate us to
define a new semantics which is free of circular justifications.

\subsection{Canonical answer sets by loop formulas}
Let $\mathcal K=(O,P)$ be a dl-program. The {\em canonical dependency graph} of $\mathcal K$,
written $G^c_\mathcal K$,
is the directed graph $(V,E)$, where $V=\HB_P$ and
$(u,v)\in E$ if there is a rule of the form (\ref{normal:rule}) in $P$ such that $A=u$ and
there exists an interpretation $I\subseteq\HB_P$ such that either
of the following two conditions holds:
\begin{enumerate}[(1)]
  \item $I\not\models_OB_i$ and $I\cup\{v\}\models_OB_i$, for some $i~(1\leq i\leq m)$. In
  this case, we say that $v$ is a {\em positive monotonic} (resp., {\em nonmonotonic})
  dependency of $B_i$ if $B_i$ is a monotonic (resp., nonmonotonic) dl-atom. Intuitively,
  the truth of $B_i$ may depend on that of $v$ while the truth of $u$ may depend on that of $B_i$. Thus
  the truth of $u$ may depend on  that of $v$.

  \item $I\models_OB_j$ and $I\cup\{v\}\not\models_OB_j$, for some $j~(1+m\leq j\leq n)$. Clearly,
  $B_j$ must be nonmonotonic.
  In this case, we say that $v$ is a {\em negative nonmonotonic dependency} of $B_j$.
  Intuitively, the truth of $u$ may depend on that of ``$\Not B_j$", while its truth
  may depend on that of $v$. Thus the truth of $u$ may depend
  on that of $v$.
\end{enumerate}
A nonempty subset $L$ of $\HB_P$ is a {\em canonical loop} of $\mathcal K$ if there is a cycle in $G_\mathcal K^c$
that goes through only and all the nodes in $L$.
It is clear  that  if $B_i=v$ then the interpretation $I=\{v\}$
satisfies $v$  while $I\setminus\{v\}$ does not. Thus the notion of canonical loops is
a generalization of that of weak loops given in Subsection~\ref{weak-loop}, and a generalization of the notion of
loops for normal logic programs \cite{LinZhao:assat}.

Note
further that the canonical dependency graph is not a generalization of the strong positive dependency graph,
since some strong loops are not canonical loops. E.g., with the dl-program
$\mathcal K=(\emptyset,P)$,
where
$P=\{p(a)\lto DL[c\odot p,c\ominus p,\neg c](a)\}$,
the dl-atom $A=DL[c\odot p, c\ominus p,\neg c](a)$ is equivalent to $\top$. So it is
monotonic. It follows that $L=\{p(a)\}$ is a
strong loop of $\mathcal K$. However $L$ is not a canonical loop of $\mathcal K$ because there
is no interpretation $I$ such that $I\not\models_OA$ and $I\cup\{p(a)\}\models_OA$.

Due to the two kinds of dependencies in a canonical dependency graph defined above,
to define canonical loop formulas, we need two kinds of irrelevant formulas:
Let $L$ be a set of atoms and $A=DL[\lambda;Q](\vec t)$ a nonmonotonic dl-atom.
The {\em positive canonical irrelevant formula} of $A$ with respect to $L$,
written $\textit{pCF}(A,L)$, is the conjunction of (1) $DL[\lambda_L;Q](\vec t)$, where
$\lambda_L$ is obtained from $\lambda$ by replacing each
predicate $p$ with $p_L$ if $L$ contains an
atom $p(\vec c)$ which is a positive nonmonotonic dependency of $A$
and, (2) for each predicate $p$ occurring in $\lambda$, the instantiation on $\mathcal C$ of the
formula (\ref{IF:formula})
if $L$ contains an atom $p(\vec c)$ which is a positive nonmonotonic dependency of $A$.
The {\em negative canonical irrelevant formula} of $A$ with respect to $L$,
written $\textit{nCF}(A,L)$, is the conjunction of (1) $DL[\lambda_L;Q](\vec t)$, where
$\lambda_L$ is obtained from $\lambda$ by replacing each
predicate $p$ with $p_L$ if $L$ contains an
atom $p(\vec c)$ which is a negative nonmonotonic dependency of $A$
and, (2) for each  predicate $p$ occurring in $\lambda$, the instantiation on $\mathcal C$ of the
formula (\ref{IF:formula})
if $L$ contains an atom $p(\vec c)$ which is a negative nonmonotonic dependency of $A$.

Let $\mathcal K=(O,P)$ be a dl-program, $M\subseteq \HB_P$ and $L$ a loop of $\mathcal K$.
The {\em canonical loop formula} of $L$ relative
to $\mathcal K$ under $M$, written $\cLF(L,M,\mathcal K)$, is the following formula:
\begin{eqnarray*}
  \bigvee L\supset\bigvee_{1\leq i\leq n}\left(\bigwedge_{A\in\Pos_i}\delta_1(A,L)\wedge\bigwedge_{B\in\Neg_i}\neg\delta_2(B,L)\right)
\end{eqnarray*}
where
\begin{itemize}
  \item $(h_1\lto\Pos_1,\Not\Neg_1),\ldots,(h_n\lto\Pos_n,\Not\Neg_n)$ are all the rules in $P$ such that $h_i\in L$,
    $\Pos_i\cap L=\emptyset$ and $M\models_O\bigwedge_{A\in\Pos_i}A\wedge\bigwedge_{B\in\Neg_i}\neg B$ for each $i~(1\leq i\leq n)$,
  \item  $\delta_1(A,L)=\textit{pCF}(A,L)$ if $A$ is a nonmonotonic dl-atom,  $\gamma(A,L)$ otherwise,
  \item $\delta_2(B,L)=\textit{nCF}(B,L)$ if $B$ is a nonmonotonic 
dl-atom, and $B$ otherwise.
\end{itemize}

Given a dl-program $\mathcal K=(O,P)$ and an interpretation $I\subseteq\HB_P$. We call $I$ a
{\em canonical answer set} of $\mathcal K$ if $I'$ is a model of $\comp(\mathcal K)\cup\cLF(I,\mathcal K)$ relative to $O$,
where $I'$ is the extension of $I$ satisfying (\ref{IF:formula}) and $\cLF(I,\mathcal K)=\{\cLF(L,I,\mathcal K)|
\mbox{$L$ is a canonical loop of $\mathcal K$}\}$. It is not difficult to prove
that every canonical answer set of a dl-program
$\mathcal K$ is a supported model of $\mathcal K$.

\begin{example}
Consider the dl-program $\mathcal K_2$ in Example \ref{exam:dl:program:1}, i.e.,
$\mathcal K_2=(\emptyset,P_2)$ where $P_2=\{p(a)\lto DL[c\oplus p, b\ominus q;c\sqcap\neg b](a)\}$.
    It is easy to see that the dl-atom $DL[c\oplus p,b\ominus q;c\sqcap \neg b](a)$ is nonmonotonic,
    $\emptyset\not\models_ODL[c\oplus p, b\ominus q; c\sqcap \neg b](a)$,
    and $\{p(a)\}\models_ODL[c\oplus p, b\ominus q; c\sqcap \neg b](a)$. Thus $L=\{p(a)\}$ is
    a canonical loop of $\mathcal K_2$. Let $I=\{p(a)\}$. The canonical loop formula $\cLF(L, I,\mathcal K)$ is
    equivalent to
    \[p(a)\supset DL[c\oplus p_L,b\ominus q;c\sqcap\neg b](a)\wedge (p_L(a)\lrto p(a)\wedge (a\neq a))\]
where the last conjunct is equivalent to $\neg p_L(a)$. Thus,
the loop formula is not satisfied by the extension of $I$ satisfying (\ref{IF:formula}) relative to the knowledge base $\emptyset$. So $I$ is not a canonical answer set of $\mathcal K_2$, even if $I$ is
a model of $\comp(\mathcal K_2)$ relative to the knowledge base $\emptyset$.
\end{example}

The next example demonstrates the difference among the positive dependency graphs
of dl-programs.
\begin{example}
 Let $\mathcal K=(O,P)$ be a dl-program where $O=\emptyset$ and $P$ consists of the following rules:\\
 \begin{tabular}{lll}
  & $p(a_1)\lto DL[c\oplus p,c](a_1)$,                       & $p(a_3)\lto \Not DL[c\ominus p,\neg c](a_3)$, \\
  & $p(a_2)\lto DL[c\oplus p,b\ominus q;c\sqcap\neg b](a_2)$,& $p(a_4)\lto p(a_4)$.
 \end{tabular}
 The only weak positive dependency on $\HB_P$ is $(p(a_4),p(a_4))$,
 the strong positive dependency includes $(p(a_1),p(a_1))$
 besides the weak one, while the canonical positive dependency contains $(p(a_2),p(a_2))$ and
 $(p(a_3),p(a_3))$ in addition to the strong ones.
\begin{figure}[t]
 \includegraphics[width=8cm]{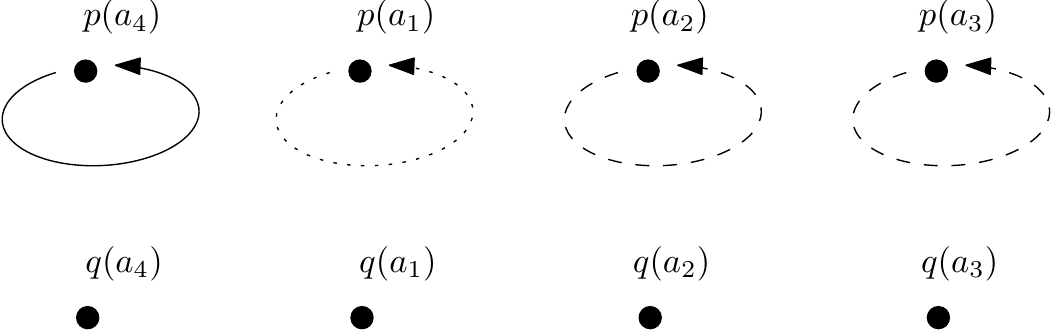}\\
 \caption{The positive dependency relations on $\HB_P$}\label{Fig:1}
\end{figure}
Figure \ref{Fig:1} depicts
the various dependency
relations on $\HB_P$.
The weak positive dependency graph is
$G_\mathcal K^w=(V,E)$ where $V=\{p(a_i),q(a_i)|1\leq i\leq 4\}$ and
 $E=\{(p(a_4),p(a_4))\}$, while the strong one is
$G_\mathcal K^s=(V,E')$ where
 $E'=E\cup\{(p(a_1),p(a_1))\}$. The canonical dependency graph is
$G_\mathcal K^c=(V,E'')$ where
 $E''=E'\cup\{(p(a_2),p(a_2)),(p(a_3),p(a_3))\}$.
\end{example}

Comparing with the previous definitions of loop formulas, in addition to
the irrelevant formulas of nonmonotonic dl-atoms, the definition of canonical loop formulas has a notable distinction: it
is given under a set $M$ of atoms whose purpose is to restrict that the support of any atom in $L$ come from the rules
whose bodies are satisfied by $M$ (relative to a knowledge base).
The next proposition shows that the canonical loops and canonical loop formulas for dl-programs
are indeed a generalization of loops and loop formulas for normal logic programs \cite{LinZhao:assat} respectively.

\begin{proposition}
  Let $P$ be a normal logic program, $L\subseteq\HB_P$ and $M$ a model of the completion of $P$.
  \begin{itemize}
    \item[(1)] $L$ is a loop of $P$ if and only if $L$ is a canonical loop of $\mathcal K=(\emptyset,P)$.
    \item[(2)] $M\models LF(L,P)$ if and only if $M\models_O\cLF(L,M,P)$, where $LF(L,P)$ is the
    loop formula associated with $L$ under $P$ \cite{LinZhao:assat} and $O=\emptyset$.
  \end{itemize}
\end{proposition}

\vspace{-.03in}
\begin{proposition}
  Let $\mathcal K=(O,P)$ be a dl-program and $I$ a canonical answer set of $\mathcal K$. Then
  $I$ is minimal in the sense that $\mathcal K$ has no canonical answer set $I'$ such that $I'\subset I$.
\end{proposition}

The following two propositions show that the canonical answer sets of dl-programs are noncircular strong answer sets.
Thus canonical answer sets are weak answer sets as well.

\vspace{-.03in}
\begin{proposition}\label{prop:CAS:non:cir}
  Let $\mathcal K=(O,P)$ be a dl-program and $I\subseteq\HB_P$ a canonical answer set of $\mathcal K$.
  Then $I$ is noncircular.
\end{proposition}

\vspace{-.03in}
\begin{proposition}
  Let $\mathcal K=(O,P)$ be a dl-program and $I\subseteq\HB_P$ a canonical answer set of $\mathcal K$.
  Then $I$ is a strong answer set of $\mathcal K$.
\end{proposition}

The following proposition, together with Proposition \ref{prop:CAS:non:cir},
implies that the operator $\ominus$ is the only cause that a strong
answer set of a dl-program is circular.

\vspace{-.03in}
\begin{proposition}
  Let $\mathcal K=(O,P)$ be a dl-program in which $P$ does not mention the operator $\ominus$.
  Then $I\subseteq\HB_P$ is a canonical answer set of $\mathcal K$ if and only if $I$ is
  a strong answer set of $\mathcal K$.
\end{proposition}

\vspace{-.3cm}
\section{Related Work}
\label{related-work}
Integrating
ASP with description
logics has attracted a great deal of attention recently.
The existing approaches
can be roughly classified into three categories.
The first is to adopt a nonmonotonic formalism that covers both
ASP and
first-order logic (if not for the latter, then extend it to the first-order case)
\cite{Motik:JACM:2010,eiter-auto-ontology2007}, where
ontologies
and rules are written in the same language, resulting in a tight coupling.
The second is a loose approach:
An ontology knowledge base and
the rules share the same constants but not the same predicates,
and the communication is
via a well-defined interface, such as
dl-atoms \cite{DBLP:journals/ai/EiterILST08}.
The third is to combine ontologies with
hybrid rules \cite{Rosati,DBLP:Rosati:KR:2006,DBLP:Bruijn:RR:2007}, where
predicates in the language of ontologies are interpreted classically, whereas those
in the language of rules are interpreted nonmonotonically.

Although each approach above has its own merits, the loose approach
possesses
some unique advantages.
In many situations, we would like to combine existing knowledge bases,
possibly
under different logics. In this
case, a notion of interface is natural and necessary.
The
loose approach seems particularly intuitive, as
it does not rely on the use
of modal operators nor on a multi-valued logic.
One notices that dl-programs
share similar characteristics with another recent interest,
{\em multi-context systems}, in which knowledge bases
of arbitrary logics communicate
through {\em bridge rules} \cite{Brewka-AAAI-07}.

However, the relationships among these different
approaches are currently not well understood.
For example, although we know how to translate
a dl-program without the nonmonotonic operator
$\ominus$ to an MKNF theory while preserving
the strong answer set semantics \cite{Motik:JACM:2010}, when
$\ominus$ is involved,
no such a translation
is known.
Similarly,
although a variant of
Quantified Equilibrium Logic (QEL)
captures the existing hybrid approaches, as shown by
\cite{DBLP:Bruijn:RR:2007}, it is not clear how one would apply
the loop formulas for logic programs with
arbitrary sentences \cite{Lee:KR08} to dl-programs, since, to the best of our
knowledge, there is no syntactic, semantics-preserving
translation from
dl-programs to logic programs
with arbitrary sentences or to QEL.

In fact,
the loop formulas for
dl-programs are more involved
than any previously known
loop formulas, due to mixing ASP
with
classical first-order logic.
This is evidenced by the
fact that weak loop formulas permit
self-supports, strong loop formulas eliminate
certain kind of self-supports,
and finally canonical loop formulas remove all self-supports.  This seems to
be a unique phenomenon that arises to dl-programs, not to
any other known extensions of ASP, including
logic programs
with arbitrary sentences.

\vspace{-.3cm}
\section{Concluding Remarks}
\label{Discussion-and-Future-Work}

In this paper, we characterized
the weak and
strong answer sets of
dl-programs by program completion and loop formulas.
Although
these loop formulas
also provide
an alternative mechanism for
computing answer sets,
building such a system
presents itself as an
interesting future work. We also proposed the canonical answer sets for dl-programs, which are
minimal and  noncircular in a formal sense. From the perspective of loop formulas,
we see a notable distinction among the weak, strong and canonical answer sets:
the canonical answer sets permit no circular justifications,  the strong answer sets permit
circular justifications involving nonmonotonic dl-atoms but not monotonic ones,
whereas the weak answer sets permit circular justifications that involve
any dl-atoms but not
atoms.

We remark that, for a given dl-program $\mathcal K=(O,P)$, to decide
if a set $M\subseteq\HB_P$ is a strong or canonical loop and to
construct the strong or canonical loop formula of $M$ are generally
quite difficult, since we have to decide the monotonicity of the
dl-atoms occurring in $P$.
The exact complexity of deciding if a set of atoms
is a strong or canonical loop is one of our ongoing studies, in
addition to the complexity of deciding if a given dl-program has a
canonical answer set.

\vspace{.1in}
\noindent
{\bf Acknowledgment}:
We thank the anonymous
reviewers for their detailed comments, which helped improve the presentation of the paper.
Yisong Wang was supported in part by NSFC
grants 90718009 and 60703095, the Fund of
Guizhou Science and Technology 2008[2119], the Fund of Education
Department of Guizhou Province 2008[011], Scientific Research
Fund for talents recruiting of Guizhou University 2007[042].
Jia-Huai You and Li Yan Yuan are partially supported by NSERC and
by the Ministry of Science and Technology of China under 863 plan.
Yi-Dong Shen is supported in part by NSFC grants 60970045 and 60721061.


\bibliographystyle{acmtrans}

\section*{Appendix: Proofs}
\setcounter{theorem}{0}
\setcounter{proposition}{0}

We first recall the operator $\gamma_\mathcal
K:\textit{HB}_P\rto \textit{HB}_P$ for a positive dl-program
$\mathcal K=(O,P)$ \cite{DBLP:journals/ai/EiterILST08}: let
$I\subseteq\textit{HB}_P$,
\begin{equation}
  \gamma_\mathcal K(I)=\{h|(h\lto\Pos)\in P\ and\ I\models_OA\mbox{ for any $A\in\Pos$}\}
\end{equation}
Since $\gamma_\mathcal K$ is monotonic, so it has the least
fix-point which is the unique least model of $\mathcal K$. Such
least fix-point can be iteratively constructed as:
\begin{itemize}
  \item $\gamma_\mathcal K^0=\emptyset$;
  \item $\gamma_\mathcal K^{n+1}=\gamma_\mathcal K(\gamma_\mathcal K^n)$.
\end{itemize}
It is clear that the least fix-point $\lfp(\gamma_\mathcal
K)=\gamma_\mathcal K^\infty$. So $I\subseteq\textit{HB}_P$ is a
strong (resp., weak) answer set of a dl-program $\mathcal K=(O,P)$ if and only if
$I=\lfp(\gamma_{\mathcal K^{s,I}})$ (resp., $I=\lfp(\gamma_{\mathcal K^{w,I}})$).

\begin{proposition}
  Let $\mathcal K=(O,P)$ be a dl-program and $I\subseteq\textit{HB}_P$. Then $I$ is a supported model of $\mathcal K$
  if and only if $I\models_O\comp(\mathcal K)$.
\end{proposition}
\begin{proof}
  The interpretation $I$ is a supported model of $\mathcal K$\\
  iff, for any $h\in I$, there exists a rule
  $(h\lto\Pos,\Not\Neg)$
  in $P$ such that  $$I\models_O\left(\bigwedge_{A\in\Pos}A\wedge\bigwedge_{B\in\Neg}\neg B\right)$$
  iff $I\models_O\comp(h, \mathcal K)$ for any $h\in I$\\
  iff $I\models_O\comp(\mathcal K)$.
\end{proof}

\begin{proposition}
  Let $\mathcal K=(O,P)$ be a dl-program and $I\subseteq\HB_P$ a strong (or weak) answer set of $\mathcal K$.
  Then $I$ is a supported model of $\mathcal K$.
\end{proposition}
\begin{proof} 
  (1) Let $I$ be a strong answer set of $\mathcal K$.
  It is sufficient to show that, for any $h\in I$,
  $I\models_O\comp(h,\mathcal K)$ by Proposition
  \ref{prop:completion:supported}. Note that \\
  $h\in I$\\
  $\Rto$ there is a dl-rule
    ($r':h\lto \Pos_1$)
   in $sP_O^I$ such that $I\models_OA$ for any $A\in\Pos_1$\\
  $\Rto$ there is a dl-rule
    ($r:h\lto\Pos_1,\Pos_2,\Not \Neg$) in $P$ such that $r'$ is obtained
    from $r$ by the strong dl-transformation, where $\Pos_2$ is a set of nonmonotonic
    dl-atoms, i.e., (i) $I\models_OB$ for any $B\in\Pos_2$, and (ii)
    $I\not\models_OB'$ for any $B'\in\Neg$\\
  $\Rto
  I\models_O\bigwedge_{A\in\Pos_1\cup\Pos_2}A\wedge\bigwedge_{B\in\Neg}\neg
  B$.

  Consequently, $I$ is a supported model of $\mathcal K$.

  (2) The proof is similar when $I$ is a weak answer set of $\mathcal
  K$.
\end{proof}

\begin{theorem}
  Let $\mathcal K=(O,P)$ be a dl-program and $I\subseteq \HB_P$. $I$ is a weak answer set of $\mathcal K$
  if and only if $I\models_O\comp(\mathcal K)\cup \textit{wLF}(\mathcal K)$, where $\textit{wLF}(\mathcal K)$
  is the set of weak loop formulas relative to $\mathcal K$.
\end{theorem}
\begin{proof}
  ($\Rto$) By Proposition \ref{prop:AS:supported:model}, we only need to show that, $I\models_O\wLF(L,\mathcal K)$
  for any weak loop $L$ of $\mathcal K$. Suppose $I\not\models_O\wLF(L,\mathcal K)$, i.e.,
  \begin{eqnarray}\label{thm:w:eq:1}
    I\models_O\bigvee L\ \ \textit{and}\ \ I\not\models_O\left(\bigwedge_{A\in\Pos}A\wedge\bigwedge_{B\in\Neg}\neg B\right)
  \end{eqnarray} for
  any rule $(h\lto\Pos,\Not\Neg)$ in $P$ such that $h\in L$ and $\Pos\cap L=\emptyset$. It implies that $I\cap L\neq\emptyset$.
  Without loss of generality, suppose $L=\{h_1,\ldots,h_k\}$ and $h_1\in I\cap L$.
  Because $I$ is a weak answer set of $\mathcal K$, $I=\lfp(\gamma_{\mathcal K^{w,I}})$.
  It follows that $h_1\in\lfp(\gamma_{\mathcal K^{w,I}})$. Let $k_1$ be the least number such that
  $h_1\in\gamma^{k_1+1}_{\mathcal K^{w,I}}$. Thus $wP^I_O$ must have a rule
  \[r_1: h_1\lto\Pos_1\]
  such that $\gamma^{k_1}_{\mathcal K^{w,I}}\models_OA$ for any $A\in\Pos_1$. Suppose $r_1$ is obtained from
  the following rule
  \[ h_1\lto\Pos_1,\textit{Adl}_1,\Not\Neg_1\]
  in $P$ by the weak dl-transformation, where $\textit{Adl}_1$ is a set of dl-atoms. Thus $I\models_O A$ for any $A\in\textit{Adl}_1$
  and $I\not\models_O B$ for any $B\in\Neg_1$. By  (\ref{thm:w:eq:1}), $\Pos_1\cap L\neq\emptyset$.
  Note that $h_1\notin\Pos_1$. Thus $(L\setminus\{h_1\})\cap \Pos_1\neq\emptyset$. Without
  loss of generality, suppose $h_2\in\Pos_1$. Similarly, there exists the least number $k_2$ such that
  $h_2\in\gamma^{k_2+1}_{\mathcal K^{w,I}}$. Using the construction, we may have a sequence $(k_1,k_2,\ldots,)$
  of natural numbers and a sequence $(h_1,h_2\ldots,)$ of atoms in $L\cap I$ such that
  \begin{itemize}
    \item $k_i$ is the smallest number such that
      $h_i\in\gamma^{k_i+1}_{\mathcal K^{w,I}}$,
    \item $(h_i\lto \Pos_i)$ is the rule in $wP_O^I$ such that $\Pos_i\subseteq \gamma^{k_i}_{\mathcal K^{w,I}}$, and
    \item $k_i<k_j$ for any $0\leq i< j$.
  \end{itemize}
  Since $I\cap L$ is finite, there must be some $i,j~(0\leq i<j)$ such that $h_i=h_j$. This implies
  that $k_i=k_j$. This is
a paradox. Thus $I\models_O \wLF(L,\mathcal K)$.

  ($\Lto$) Firstly, we show $I\subseteq\lfp(\gamma_{\mathcal K^{w,I}})$.
  Let $\Gamma$ be the set of rules in $wP^I_O$ whose bodies are satisfied by $I$.
  Since $I$ is a supported model of $\mathcal K$, the heads of rules in $\Gamma$ are also satisfied by $I$. Moreover,
  $I$ is the set of atoms occurring in $\Gamma$.
  Let $I^*=\lfp(\gamma_{\mathcal K_\Gamma^{w,I}})$, where $\mathcal K_\Gamma=(O,\Gamma)$.
  Let $I^{-}=I\setminus I^*$ and $\Gamma_{I^{-}}$ be the set
  of rules in $\Gamma$ whose heads are in $I^{-}$. We show that $(O,\Gamma_{I^{-}})$ has at least one terminating loop.

  For any rule $(r:h\lto\Pos)$ in $\Gamma_{I^{-}}$, $\Pos\subseteq I$ since $I\models A$ for any $A\in\Pos$ and
  $wP^I_O$ mentions only atoms. However $\Pos\setminus I^*\neq\emptyset$ otherwise $I^*\models A$ for
  any $A\in \Pos$ and then $r\notin\Gamma_{I^{-}}$. It implies that \[\Pos\cap (I\setminus I^*)\neq\emptyset.\]
  Suppose $h'\in\Pos\cap I^{-}$. Then there is an edge $(h,h')$ in the weak positive dependency graph of $(O,\Gamma_{I^{-}})$.
  So we can construct a sequence of atoms $$(h_1,h_2,\ldots,h_i,\ldots)$$ such that $h_i\in I^{-}$ for any $i\ge 1$ and
  $(h_i,h_{i+1})$ is an edge of the weak positive dependency graph of $(O,\Gamma_{I^{-}})$. Since $I^{-}$ is finite, the
  above sequence must contain a loop. It is clear that  if a graph has a loop then it has at least one terminating loop.
  Now suppose $L=\{h_1,\ldots, h_k\}$ is a terminating loop of $\Gamma_{I^{-}}$. We further claim that, for any rule
  $(h\lto\Pos)$ in $\Gamma_{I^{-}}$ such that $h\in L$:
  \[I^{-}\cap \Pos\subseteq L.\]
  Otherwise, we can construct
  another path $(h,h',\ldots)$ in the positive weak dependency graph of $(O,\Gamma_{I^{-}})$ such that $h'\in I\cap \Pos$ and $h'\notin L$.
  Thus we have a path from $L$ to another maximal loop of
  the weak dependency graph of $(O,\Gamma_{I^{-}})$, which contradicts
the fact that $L$ is a terminating loop.

  Note that $L$ is also a weak loop of $\mathcal K$, $I\models_O\wLF(L,\mathcal K)$ and $L\subseteq I^{-}$. It follows that
  $P$ should have at least one rule
  \[r': h'\lto\Pos',\Not\Neg'\]
  such that $\Pos'\cap L=\emptyset$, $I\models_O A$ for any $A\in\Pos'$ and $I\not\models_OB$ for any $B\in\Neg'$,
  where $h'\in L$. Suppose $(r^*: h'\lto\Pos^*)$ is the rule obtained from $r'$ by the weak dl-transformation.
  Evidently, $r^*\in\Gamma$. Furthermore $r^*\in\Gamma_{I^{-}}$ since $h'\in L\subseteq I^{-}$.
  This implies that $I^{-}\cap\Pos^*\subseteq L$
  which contradicts with $\Pos^*\cap L=\emptyset$ since $I^{-}\cap\Pos^*\neq\emptyset$.
  Consequently, $I^{-}=\emptyset$ and then $I\subseteq I^*=\lfp(\gamma_{\mathcal K_\Gamma^{w,I}})\subseteq I$.
  It implies that $I=I^*$ and
  $I\subseteq\lfp(\gamma_{\mathcal K^{w,I}})$ by $\Gamma\subseteq wP_O^I$.

  Secondly, we prove $\lfp(\gamma_{\mathcal K^{w,I}})\subseteq I$. Let $I'=\lfp(\gamma_{\mathcal K^{w,I}})\setminus I$.
  Suppose $I'\neq\emptyset$. Let $h$ be an arbitrary atom in $I'$. There is the least number $k$ such that
  $h\in\gamma^{k+1}_{\mathcal K^{w,I}}$. So that there exists a rule $(r':h\lto\Pos)$ in $wP_O^I$ such that
  $\Pos\subseteq\gamma^k_{\mathcal K^{w,I}}$. Note that $h\notin I$ and $I\models_O\comp(h,\mathcal K)$.
  It follows that, for any rule $(h\lto\Pos,\Not \Neg)$ in $P$,
  \[I\not\models_O\bigwedge_{A\in\Pos}A\wedge\bigwedge_{B\in\Neg}\neg B.\]
  It implies that $\Pos\not\subseteq I$. Thus there exists an atom $h'\in\Pos$ such that
  $h\in \gamma^k_{\mathcal K^{w,I}}\setminus I$. So we can construct a sequence of numbers
  $(k_0,k_1,\ldots)$ and a sequence $(h_1,h_2,\ldots)$ of atoms in $I'$ such that, for any $i\ge 0$,
  \begin{itemize}
    \item $k_i$ is the least number such that $h_i\in\gamma^{k_i+1}_{\mathcal K^{w,I}}$,
    \item $h_i\lto\Pos_i$ is the rule in $wP_O^I$ such that $\Pos_i\subseteq\gamma^{k_i}_{\mathcal K^{w,I}}$, and
    \item $k_i>k_j$ for any $0\leq i<j$.
  \end{itemize}
  Since $I'$ is finite, there exists $0\leq i<j$ such that $h_i=h_j$ which implies that $k_i=k_j$. It
  contradicts with $k_k>k_j$. Thus $I'=\emptyset$, i.e., $\lfp(\gamma_{\mathcal K^{w,I}})\subseteq I$.

  Consequently $I$ is a weak answer set of $\mathcal K$.
\end{proof}

\begin{lemma}\label{lem:I:I'}
  Let $\mathcal K=(O,P)$ be a dl-program, $I\subseteq\textit{HB}_P$, $I'$ is the extension
  of $I$ satisfying (\ref{IF:formula}) and $L$ be an arbitrary nonempty set of atoms. Then we have,
  for any dl-atom $A$, $I'\models_O\textit{IF}(A,L)$ iff $I\setminus L\models_OA$.
\end{lemma}
\begin{proof}
  Since $I'$ is the extension of $I$ satisfying (\ref{IF:formula}),
  we have that $p(\vec c)\in I$ iff $p(\vec c)\in I'$ for any $p(\vec c)\in\HB_P$. Furthermore,
  for any atom $p_L(\vec c)$, $p_L(\vec c)\in I'$  iff $p(\vec c)\in I\setminus L$.
  Without loss of generality, let $A=DL[S\oplus p,S'\ominus q;Q](\vec t)$. It obviously holds that
  if the predicates $p$ and $q$ do not occur in $L$ since $\textit{IF}(A,L)=A$. Let's assume that the predicates
  $p$ and $q$ appear in $L$. \\
  $I'\models_O \textit{IF}(A,L)$\\
  $\LRto I'\models_O DL[S\oplus p_L,S'\ominus q_L;Q](\vec t)$\\
    $\LRto$ $O\cup\{S(\vec e)|p_L(\vec e)\in I'\}\cup\{\neg S'(\vec e)|q_L(\vec e)\notin I'\}\models Q(\vec t)$\\
    $\LRto$ $O\cup\{S(\vec e)|p(\vec e)\in I\setminus L\}\cup\{\neg S'(\vec e)|q(\vec e)\notin I\setminus L\}\models Q(\vec t)$\\
    $\LRto I\setminus L\models_ODL[S\oplus p,S'\ominus q;Q](\vec t)$\\
    $\LRto$ $I\setminus L\models_O A$.

  The other two cases, namely (i) $p$ appears in $L$ but not $q$, and (ii) $q$ appears in $L$ but not $p$, can be similarly proved.
\end{proof}

\begin{lemma}\label{lem:sLF:1}
  Let $\mathcal K=(O,P)$ be a dl-program and $I\subseteq\HB_P$ such that
  $I\models_O\comp(\mathcal K)$. Then we have that $\lfp(\gamma_{\mathcal
  K^{s,I}})\subseteq I$.
\end{lemma}
\begin{proof}
  Let $I'=\lfp(\gamma_{\mathcal K^{s,I}})$ and $I^{-}=I'\setminus I$.
  If $I'\not\subseteq I$ then $I^{-}\neq\emptyset$. Suppose $I'\not\subseteq I$.
  For any $h\in I^{-}$, there exists the least natural number $k$ and a rule $(r:h\lto\Pos)$ in $sP_O^I$
  such that $\gamma^k_{\mathcal K^{s,I}}\models_OA$ for any $A\in\Pos$. But we know that $h\notin I$ and
  $I\models_O\comp(h,\mathcal K)$ which implies that, for any rule $(r':h\lto\Pos',\Not\Neg')$ in $P$:
  \[I\not\models_O\bigwedge_{A\in\Pos'}A\wedge\bigwedge_{B\in\Neg'}\neg B.\]
  It follows that $I\not\models_OA$ for some $A\in\Pos$. It implies that either
  \begin{enumerate}[(i)]
    \item there is some atom $h'\in \Pos\cap\gamma^k_{\mathcal K^{s,I}}$ such that $h'\notin I$, or
    \item there is a monotonic dl-atom $A=DL[\lambda;Q](\vec t)$ in $\Pos$ such that, for some
    $S\oplus p$ (or $S\odot p$) occurring in $\lambda$, there is an atom $h'=p(\vec c)$ such that
    $h'\in\gamma^k_{\mathcal K^{s,I}}$
    and $p(\vec c)\notin I$.
  \end{enumerate}
  It is evident that $h'\neq h$ and $h'\in I^{-}$. Thus we have a sequence $(k_0,k_1,\ldots)$ of natural numbers
  and a sequence $(h_1,h_2,\ldots)$ of atoms in $I^{-}$ such that: for any $i\ge 0$,
  \begin{itemize}
    \item $k_i$ is the least number such that $h_i\in\gamma^{k_i+1}_{\mathcal K^{s,I}}$,
    \item $(h_i\lto\Pos_i)$ is in $sP_O^I$ such that $\gamma^{k_i}_{\mathcal K^{s,I}}\models_OA$ for any $A\in\Pos_i$, and
    \item $k_i>k_j$ for any $0\leq i <j$.
  \end{itemize}
  Since $I^{-}$ is finite, in the above sequence of atoms there must be $i,j~(0\leq i<j)$ such that $h_i=h_j$.
  It implies that $k_i=k_j$ which contradicts with $k_i>k_j$. Consequently, $I^{-}=\emptyset$, i.e., $I'\subseteq I$.
\end{proof}

\begin{theorem}
  Let $\mathcal K=(O,P)$ be a dl-program and $I\subseteq \textit{HB}_P$. $I$ is a strong answer set of $\mathcal K$
  if and only if $I'\models_O\comp(\mathcal K)\cup \sLF(\mathcal K)$, where $\sLF(\mathcal K)$ is the set
  of strong loop formulas of all strong loops of $\mathcal K$ and $I'$ is the extension of $I$
  satisfying (\ref{IF:formula}).
\end{theorem}
\begin{proof}
It is clear that $I'\cap\HB_P=I$ since $I'$ is the extension of $I$ satisfying (\ref{IF:formula}).

  ($\Rto$) Evidently, $I'\models_O\comp(\mathcal K)$. By Proposition \ref{prop:AS:supported:model},
  it is sufficient
   to prove that, for any strong loop $L$ of $\mathcal K$, $I'\models_O \sLF(L,\mathcal K)$.
    Suppose $L=\{h_1,\ldots,h_k\}$ is a strong loop of $\mathcal K$ and $I'\not\models_O\sLF(L,\mathcal K)$, i.e.,
    \[I'\models_O\bigvee L\ \ \ \textit{and}\ \ \ I'\not\models_O
        \bigvee_{1\leq i\leq n}\left(\bigwedge_{A\in\Pos_i}\gamma(A,L)\wedge\bigwedge_{B\in\Neg_i}\neg B\right)\]
    where $(h_1\lto \Pos_1,\Not\Neg_1),\ldots, (h_n\lto\Pos_n,\Not\Neg_n)$ are all the rules in $P$  such that
    $h_i\in L$ and $\Pos_i\cap L=\emptyset$ for any $i~(1\leq i\leq n)$. It follows that, for any $i~(1\leq i\leq n)$,
    \begin{eqnarray}\label{thm:eq:1}
        I'\not\models_O\bigwedge_{A\in\Pos_i}\gamma(A,L)\wedge\bigwedge_{B\in\Neg_i}\neg B.
    \end{eqnarray}
    Since $I'\models_O \bigvee L$, we have that $I'\cap L\neq\emptyset$ and then $I\cap L\neq\emptyset$.
    Without loss of generality, let's assume
    $h_1\in I\cap L$. Note that $I$ is a strong answer set of $\mathcal K$, i.e., $I=\lfp(\gamma_{\mathcal K^{s,I}})$.
    Thus there is the least number $k_1$
    such that $h_1\in\gamma^{k_1+1}_{\mathcal K^{s,I}}$.
    So there is a rule ($r_1:h_1\lto \Pos_1$)
    in $sP^I_O$ such that $\gamma^{k_1}_{\mathcal K^{s,I}}\models_OA$ for any $A\in\Pos_1$.
    It is evident that $h_1\notin\Pos_1$. It implies that $P$ has
    a rule $$r'_1:h_1\lto \Pos_1,\textit{Ndl}_1,\Not \Neg_1,$$
    where $\textit{Ndl}$ is a set of nonmonotonic dl-atoms, such that
    $r_1$ is obtained from $r_1'$ by the strong dl-transformation, i.e.,   $I\models_OA$ for any
    $A\in\textit{Ndl}_1$ and
    $I\not\models_OB$ for any $B\in\Neg_1$. Note that $I'\cap\HB_P=I$.  It
    is clear that, $I'\models_OA$ for each $A\in \textit{Ndl}_1$ and $I'\not\models_O B$
    for any $B\in\Neg_1$. By  (\ref{thm:eq:1}), at least one of the
    following two cases holds:
    \begin{itemize}
        \item $\Pos_1\cap L\neq \emptyset$. In this case, there is some atom $h\in\Pos_1\cap L$
        and $h\neq h_1$.

        \item $I'\not\models_O \textit{IF}(A,L)$ for some monotonic dl-atom
        $A=\textit{DL}[\lambda;Q](\vec t)$ in $\Pos_1$. By Lemma \ref{lem:I:I'}, we have $I\setminus
        L\not\models_OA$. Since $A$ is monotonic, then we further have
        $\gamma^{k_1}_{\mathcal K^{s,I}}\setminus L\not\models_OA$. But we know that
        $\gamma^{k_1}_{\mathcal K^{s,I}}\models_OA$.
        It follows that, there exists some atom $p(\vec c)\in L\cap\gamma^{k_1}_{\mathcal K^{s,I}}$,
        $p(\vec c)\neq h_1$ and $S\oplus p$ (or $S\odot p$) appears in $L$ for some $S$.
    \end{itemize}

    By the above analysis, we can have a sequence of natural numbers $(k_1,k_2,\ldots,)$
    and a sequence $(h_1,h_2,\ldots)$ of atoms in $L$ such that, for any $i\ge 1$,
    \begin{itemize}
      \item $k_i$ is the least natural number such that $h_i\in\gamma^{k_i+1}_{\mathcal K^{s,I}}$,
      \item $(h_i\lto\Pos_i)$ is the rule in $sP_O^I$ such that $\gamma^{k_i+1}_{\mathcal K^{s,I}}\models_OA$ for
      any $A\in\Pos_i$, and
      \item $k_i>k_j$ for any $1\leq i<j$.
    \end{itemize}

    Since $L$ is finite, there must be some $i,j~(1\leq i<j)$ such that $h_i=h_j$, which
    implies that $k_i=k_j$. This is a paradox.
    Consequently, $I'\models_O \sLF(L,\mathcal K)$.

   ($\Lto$) Let $I=I'\cap\HB_P$. By
Proposition~\ref{prop:completion:supported}, $I$ is a supported
model of $\mathcal K$.
   Let $\Gamma$ be the set of rules in $sP^I_O$ whose bodies are satisfied by $I$ relative to $O$.
   Clearly, for any rule $(h\lto \Pos)$ in $\Gamma$, $h\in I$. And
   inversely, for any $h\in I$, there exists at least one rule
   $(h\lto\Pos)$ in $\Gamma$.
   Let $I^*=\lfp(\gamma_{\mathcal K_\Gamma^{s,I}})$ where $\mathcal K_\Gamma=(O,\Gamma)$.
   Evidently, $I^*\subseteq I$. Let
   $I^{-}=I\setminus I^*$. Suppose $I^{-}\neq\emptyset$.
   Let $\Gamma_{I^{-}}$ be the set of rules in $\Gamma$ whose heads belong to $I^{-}$. We claim that
   the dl-program $(O,\Gamma_{I^{-}})$
   must have one terminating loop.

   Firstly, let $h\in I^{-}$ and suppose $(h\lto\Pos)$ be a rule in $\Gamma_{I^{-}}$. We have that
   \[I^*\not\models_O\bigwedge_{A\in\Pos}A\ \ \textit{and}\ \ I^*\cup I^{-}\models_O\bigwedge_{A\in\Pos}A.\]
   It follows that there is an atom or dl-atom $A$ in $\Pos$ such that $I^*\not\models_OA$.
   That implies that at least one of the following cases hold:
   \begin{itemize}
     \item there is some atom $h'\in\Pos$, $h'\in I^{-}$;
     \item there exists a monotonic dl-atom $A=DL[\lambda;Q](\vec t)$ in $\Pos$ such that $I^*\not\models_OA$,
     which implies that there exists some $S\oplus p$ (or $S\odot p$) appearing in $\lambda$
     and $p(\vec c)\in I\setminus I^*$ for some atom $p(\vec c)$ since $I\models_OA$,
     otherwise $I^*\models_OA$.
   \end{itemize}

   It follows that, there exists an edge $(h,h')$ in the positive strong dependency graph $G$ of
   the dl-program $(O,\Gamma_{I^{-}})$
   where $h'\in I^{-}$. Consequently, we can construct a sequence \[(h_0,h_1,\ldots,h_i,\ldots)\] of atoms
   in $I^{-}$ such that, for any $i\ge 0$, $(h_i,h_{i+1})$ is an edge of $G$. Since $I^{-}$ is finite, the constructed
   sequence must contain a loop. Furthermore, $G$ has at least one terminating loop.
   Let $L$ be a terminating loop of $(O,\Gamma_{I^{-}})$, $h\in L$ and
   \begin{eqnarray*}\label{thm:rule:1}
    r:h\lto\Pos
   \end{eqnarray*}
   be an arbitrary rule
   in $\Gamma_{I^{-}}$. It is obvious that $L\subseteq I^{-}$. Because $L$ is a terminating loop of $(O,\Gamma_{I^{-}})$,
   it follows that the following cases hold:
   \begin{itemize}
     \item $I^{-}\cap \Pos\subseteq L$, and
     \item for any monotonic dl-atom $DL[\lambda;Q](\vec t)$ in $\Pos$,
     if $S\oplus p$ (or $S\odot p$) appear in $\lambda$ for some $S$ then we have $p(\vec c)\in I^{-}$ implies $p(\vec c)\in L$.
   \end{itemize}

   Note that $L$ is also a loop of $\mathcal K$. Due to $I'\models_O \sLF(L,\mathcal K)$, $L\subseteq I^{-}$,
   and $I^{-}\subseteq I'$,
   we have $I'\models_O \bigvee L$. Thus, $P$ has at least
   one rule
   \begin{eqnarray*}
     r':h'\lto \Pos',\Not\Neg'
   \end{eqnarray*}
   such that $h'\in L$, $\Pos'\cap L=\emptyset$ and
   \[I'\models_O \left(\bigwedge_{A\in\Pos'}\gamma(A,L)\wedge\bigwedge_{B\in\Neg'}\neg B\right).\]
   It implies that $I\models_OA$ for any nonmonotonic dl-atom $A\in\Pos'$ and $I\not\models_OB$
   for any $B\in\Neg'$. Let $(r'':h'\lto\Pos'')$ be the rule in $sP^I_O$ that is obtained from $r'$ by
   the strong dl-transformation. Clearly, $r''\in\Gamma$ by Lemma \ref{lem:I:I'}.
   Furthermore, due to $h'\in L\subseteq I^{-}$, so we have
   \begin{eqnarray*}
     r''\in \Gamma_{I^{-}}.
   \end{eqnarray*}
   Note that $\Pos''\cap I^{-}\subseteq L$ implies $\Pos'\cap I^{-}\subseteq L$.
   It follows that $\Pos'\cap I^{-}=\emptyset$  by $\Pos'\cap L=\emptyset$. So we have
   $\Pos'\cap\HB_P\subseteq I^*$ since $I\models A$ for any $A\in\Pos'\cap\HB_P$.
   It implies that $\Pos''\cap \HB_P\subseteq I^*$.
   Since $r''\in\Gamma_{I^{-}}$, $\Pos''$ must have a monotonic dl-atom $A=DL[\lambda;Q](\vec t)$
   such that $I^*\not\models_OA$, i.e., $I\setminus I^{-}\not\models_OA$.
   By Lemma \ref{lem:I:I'}, we have $I\setminus L\models_O A$
   since $I'\models_O\gamma(A,L)$. Thus there must exist some atom $p(\vec c)\in
   (I\setminus L)\setminus(I\setminus I^{-})(=I^{-}\setminus L)$
   and $S\oplus p$ (or $S\odot p$) appears in $\lambda$ since $A$ is monotonic.
   However, we know that, for any such above atom $p(\vec c)$, $p(\vec c)\in I^{-}$ implies $p(\vec c)\in L$.
   It follows that $I\setminus L\not\models_OA$ by $I\setminus I^{-}\not\models_OA$. It is a paradox.

   Consequently, $I\setminus I^{-}=\emptyset$. It implies that $I\subseteq I^*=\lfp(\gamma_{\mathcal K_\Gamma^{s,I}})$.
   Note that $\Gamma\subseteq sP_O^I$. We have that
   $\lfp(\gamma_{\mathcal K_\Gamma^{s,I}})\subseteq \lfp(\gamma_{\mathcal K^{s,I}})$.
   It follows that $I\subseteq\lfp(\gamma_{\mathcal K^{s,I}})$. By Lemma \ref{lem:sLF:1},
   $\lfp(\gamma_{\mathcal K^{s,I}})\subseteq I$ since $I\models_O\comp(\mathcal K)$.
   Consequently, $I=\lfp(\gamma_{\mathcal K^{s,I}})$. Thus $I$ is a strong answer set of $\mathcal K$.
\end{proof}

\begin{proposition}
  Let $\mathcal K=(O,P)$ be a dl-program, $I$ an interpretation of $P$ and $L$ a weak loop of $\mathcal K$. Then
  we have $I'\models_O\sLF(L,\mathcal K)\supset\wLF(L,\mathcal K)$,
  where $I'$ is the extension of $I$ according to (\ref{IF:formula}).
\end{proposition}
\begin{proof}
  Suppose $I'\models_O\sLF(L,\mathcal K)$ and $I'\not\models_O\wLF(L,\mathcal K)$. We have that
  $I'\cap L\neq\emptyset$ and, for any dl-rule $(h\lto\Pos,\Not\Neg)$ in
  $P$ such that $h\in L$ and $\Pos\cap L=\emptyset$,
  \[I'\not\models_O \bigwedge_{A\in\Pos}A\wedge\bigwedge_{B\in\Neg}\neg B.\]
  Note that $L$ is also a strong loop of $\mathcal K$ and $I'\models_O\sLF(L,\mathcal K)$. It implies
  that there exists at least one rule $(h'\lto\Pos',\Not\Neg')$ in $P$ such that
  $h'\in L$, $\Pos'\cap L=\emptyset$ and
  \[I'\models_O\bigwedge_{A'\in\Pos'}\gamma(A',L)\wedge\bigwedge_{B'\in\Neg'}\neg B'.\]
  It is clear that, for any formula $\psi$ of $\mathcal L_\mathcal K$, $I'\models_O\psi$
  implies that $I\models_O\psi$ since $\psi$ mentions only the predicates occurring in $\mathcal K$.
  Notice further that if $A'$ is a monotonic dl-atom then $I\models_OA$ by Lemma \ref{lem:I:I'}.
  It follows that \[I\models_O\bigwedge_{A'\in\Pos'}A'\wedge\bigwedge_{B'\in\Neg'}\neg B'\]
  which contradicts with $I\not\models_O\wLF(L,\mathcal K)$.
\end{proof}

\begin{lemma}\label{lem:cano:dep}
  Let $\mathcal K=(O,P)$ be a dl-program, $A$ be a dl-atom appearing in $P$,
  $I_1\subset I_2\subseteq\HB_P$.
  \begin{enumerate}[(1)]
    \item If $I_1\not\models_OA$ and $I_2\models_OA$
  then there exists an interpretation $I^*$ and an atom $h^*\in I_2\setminus I_1$ such that
  $I_1\subset I^*\subseteq I_2$,
  $I^*\models_OA$ and $I^*\setminus\{h^*\}\not\models_OA$.
    \item If $A$ is nonmonotonic, $I_1\models_OA$ and $I_2\not\models_OA$
  then there exists an interpretation $I^*$ and an atom $h^*\in I_2\setminus I_1$ such that $I_1\subseteq I^*\subset I_2$,
  $I^*\cup\{h^*\}\not\models_OA$ and $I^*\models_OA$.
  \end{enumerate}
\end{lemma}
\begin{proof}
  (1) It is clear that $I_2\setminus I_1\neq\emptyset$ by the assumption. We construct an interpretation $I^*$ by
  Algorithm  \ref{algoth:1}.

\begin{algorithm}[t]
 \caption{\emph{Psup}($A,I_1,I_2)$}
 \begin{algorithmic}

  \STATE $I^*\lto I_2$
  \STATE $M\lto I_2\setminus I_1$
  \FORALL{ $h\in M\cap I^*$ }
        \STATE $h^*\lto h$
        \IF{$I^*\setminus\{h^*\}\models_OA$}
            \STATE $I^*\lto I^*\setminus \{h^*\}$
            \STATE \textbf{continue}
        \ENDIF
       \STATE \textbf{break}
  \ENDFOR
  \RETURN $(I^*,h^*)$
 \end{algorithmic}
  \label{algoth:1}
\end{algorithm}

\comment{
  \SetKwInOut{Input}{input}
  \SetKwInOut{Output}{output}
  \restylealgo{boxed}
  \linesnumbered
  \begin{algorithm}
  \Input{$A,I_1$ and $I_2$}
  \Output{$I^*$ and $h^*$}
  \Begin{
    $I^*\lto I_2$\;
    $M\lto I_2\setminus I_1$\;
    \ForEach{ $h\in M\cap I^*$ }{
        $h^*\lto h$\;
        \If{$I^*\setminus\{h^*\}\models_OA$}{
            $I^*\lto I^*\setminus \{h^*\}$;
            continue\;
        }
        break;
    }
    \Return $(I^*,h^*)$;
  }
  \caption{\emph{Psup}($A,I_1,I_2)$}
  \end{algorithm}}
  Since both $I_2$ and $M$ are finite, the algorithm definitely terminates. Note that
  $M$ is a nonempty subset of $I_2$, the \textbf{forall} loop will run at least once. Suppose
  {\em Psup($A,I_1,I_2)$} is terminated.
  There are only two cases leading to its termination:
  \begin{itemize}
    \item There is no $h\in M\cap I^*$ (line 3). It implies that $I^*=I_1$ and $I^*\models_OA$.
    The latter contradicts with $I_1\not\models_OA$. Thus this case is impossible.
    \item The ``break" is executed (line 9). It implies that $I^*\subseteq I_2$ and
    $I^*\setminus \{h^*\}\not\models_OA$.
  \end{itemize}
  Thus the above algorithm returns ($I^*, h^*$) satisfying the condition
  $I^*\models_OA$ and $I^*\setminus\{h^*\}\not\models_OA$.

  (2) We have Algorithm \ref{algo:2} for this purpose.
  \begin{algorithm}[t]
    \caption{\emph{Nsup}($A,I_1,I_2$)}
    \begin{algorithmic}
        \STATE $I^*\lto I_1$
        \STATE $M\lto I_2\setminus I_1$
        \FORALL{ $h\in M\setminus I^*$ }
            \IF{$I^*\cup\{h\}\models_OA$}
                \STATE $I^*\lto I^*\cup \{h\}$
                \STATE \textbf{continue}
            \ENDIF
            \STATE $h^*\lto h$
            \STATE $I^*\lto I^*\cup\{h^*\}$
            \STATE \textbf{break}
        \ENDFOR
        \RETURN $(I^*,h^*)$
    \end{algorithmic}
    \label{algo:2}
  \end{algorithm}
\comment{
 \SetKwInOut{Input}{input}
  \SetKwInOut{Output}{output}
  \restylealgo{boxed}
  \linesnumbered
  \begin{algorithm}
  \Input{$A,I_1$ and $I_2$}
  \Output{$I^*$ and $h^*$}
  \Begin{
    $I^*\lto I_1$\;
    $M\lto I_2\setminus I_1$\;
    \ForEach{ $h\in M\setminus I^*$ }{
        \If{$I^*\cup\{h\}\models_OA$}{
            $I^*\lto I^*\cup \{h\}$;
            continue\;
        }
        $h^*\lto h$\;
        $I^*\lto I^*\cup\{h^*\}$\;
        break;
    }
    \Return $(I^*,h^*)$\;
  }
  \caption{\emph{Nsup}($A,I_1,I_2$)}
  \end{algorithm}}
  Similarly, since both $M$ and $I_2$ are finite then the algorithm {\em Nsup} definitely terminates
  and the \textbf{forall} loop will be executed at least once. Suppose {\em Nsup$(A,I_1,I_2)$}
  is executed and terminated. If {\em Nsup} terminates because of
  $M\setminus I^*=\emptyset$ in the \textbf{forall} loop, in this case, we have $I^*=I_2$ and
  $I^*\models_OA$. The latter contradicts with $I_2\not\models_OA$. Thus the only case leading
  to the termination of {\em Nsup} is the ``break" (line 10). In that case, we have that $I^*\cup\{h^*\}\not\models_OA$
  and $I^*\models_OA$. It is obvious  $I_1\subseteq I^*$.
\end{proof}

\begin{lemma}\label{lem:cDF}
  Let $\mathcal K=(O,P)$ be a dl-program, $I\subseteq\HB_P$, $L$ a set of atoms and $A=DL[\lambda;Q](\vec t)$
  a nonmonotonic dl-atom appearing in $P$.
  \begin{itemize}
    \item[(1)] If $I'\models_O\textit{pCF}(A,L)$ then
     $I\setminus L\models_OA$,
    \item[(2)] If $I'\not\models_O\textit{nCF}(A,L)$ then
     $I\setminus L\not\models_OA$,
  \end{itemize} where
  $I'$ is the extension of $I$ according to (\ref{IF:formula}).
\end{lemma}
\begin{proof}
  Without loss of generality, let $\lambda=(S_1\oplus p_1,S_2\ominus p_2)$  for clarity.

  (1) Suppose $p_1\neq p_2$. There is no atom $p_2(\vec c)$ which
  is a positive nonmonotonic dependency of $A$. If there is no atom $p_1(\vec c)\in L$ such that $p_1(\vec c)$ is a positive nonmonotonic
  dependency of $A$ then $\textit{pCF}(A,L)=A$. It follows that $I\models_OA$ since $I'\models_OA$
  and $I'$ is the extension of $I$. Suppose $I\setminus L\not\models_OA$. From (1) of Lemma \ref{lem:cano:dep},
  there is an atom $h\in I\setminus (I\setminus L)$, i.e., $h\in L$, and an interpretation $I^*$
  such that $I^*\not\models_OA$ and $I^*\cup \{h\}\models_OA$.  It is evident that $h$
  must mention the predicate $p_1$. It implies that $h$ is a positive
  nonmonotonic dependency of $A$ which contradicts with the assumption.  Thus $I\setminus L\models_OA$.

  Suppose there is some atom $p_1(\vec c)\in L$ such that $p_1(\vec c)$ is a positive nonmonotonic
  dependency of $A$. Note that $p_{1_L}(\vec c)\in I'$ iff $p_1(\vec c)\in I\setminus L$ according to
  (\ref{IF:formula}).  \\
  $I'\models_O\textit{pCF}(A,L)$\\
  $\Rto O\cup\{S_1(\vec e)|p_{1_L}(\vec c)\in I'\}\cup\{\neg S_2(\vec e)|p_2(\vec e)\notin I'\}\models Q(\vec t)$\\
  $\Rto O\cup\{S_1(\vec e)|p_1(\vec c)\in I\setminus L\}\cup\{\neg S_2(\vec e)|p_2(\vec e)\notin I\}\models Q(\vec t)$\\
  $\Rto O\cup \{S_1(\vec e)|p_1(\vec c)\in I\setminus L\}\cup\{\neg S_2(\vec e)|p_2(\vec e)\notin I\setminus L\}\models Q(\vec t)$\\
  $\Rto I\setminus L\models_O DL[S_1\oplus p_1,S_2\ominus p_2;Q](\vec t)$\\
  $\Rto I\setminus L\models_OA$.

  It is similar to show that $I\setminus L\models_OA$ for the case $p_1=p_2$.

  (2) Suppose $p_1\neq p_2$. There is no atom $p_1(\vec c)$ which is a negative nonmonotonic dependency
  of $A$. If there is no atom $p_2(\vec c)\in L$ such that $p_2(\vec c)$ is a negative
  nonmonotonic dependency of $A$ then $\textit{nCF}(A,L)=A$. It implies that $I\not\models_OA$
  since $I'\not\models_OA$ and $I'$ is the extension of $I$. Suppose $I\setminus L\models_OA$.
  By (2) of Lemma \ref{lem:cano:dep}, there is some atom $h\in I\setminus (I\setminus L)$,
  i.e., $h\in L$, and an interpretation $I^*$ such that $I^*\models_OA$ and $I^*\cup\{h\}\not\models_OA$.
  It is clear that $h$ must mention the predicate $p_2$. It implies that $h$ is a negative nonmonotonic
  dependency of $A$ which contradicts with the assumption. Thus $I\setminus L\not\models_OA$.

  Suppose there is some atom $p_2(\vec c)\in L$ such that $p_2(\vec c)$ is a negative nonmonotonic
  dependency of $A$. Note that $p_{2_L}(\vec c)\in I'$ iff $p_2(\vec c)\in I\setminus L$ according to
  (\ref{IF:formula}).\\
  $I'\not\models\textit{nCF}(A,L$)\\
  $\Rto O\cup\{S_1(\vec e)|p_1(\vec e)\in I'\}\cup\{\neg S_2(\vec e)|p_{2_L}(\vec e)\not\in I'\}\not\models Q(\vec t)$\\
  $\Rto O\cup\{S_1(\vec e)|p_1(\vec e)\in I\}\cup\{\neg S_2(\vec e)|p_2(\vec e)\not\in I\setminus L\}\not\models Q(\vec t)$\\
  $\Rto O\cup\{S_1(\vec e)|p_1(\vec e)\in I\setminus L\}\cup\{\neg S_2(\vec e)|p_2(\vec e)\not\in I\setminus L\}\not\models Q(\vec t)$\\
  $\Rto I\setminus L\not\models_ODL[S_1\oplus p_1,S_2\ominus p_2;Q](\vec t)$\\
  $\Rto I\setminus L\not\models_OA$.

  It is similar to show that $I\setminus L\not\models_OA$ for the case $p_1=p_2$.
\end{proof}

Please note that the inverses of (1) and (2) do not generally hold. For example,
let $A=DL[S_1\oplus p_1,S_2\ominus p_2;S_1\sqcap\neg S_2](a)$, $I_1=\{p_1(a),p_2(a)\}$, $I_2=\{p_1(a)\}$,
$L_1=\{p_2(a)\}$ and $L_2=\{p_1(a)\}$. Because there is no interpretation $I$
such that $I\not\models_OA$ and $I\cup L_1\models_OA$, it implies $\textit{pCF}(A,L_1)=A$.
Similarly, we have $\textit{nCF}(A,L_2)=A$. Note that $I_1\setminus L_1\models_OA$. However
$I'_1\not\models_O\textit{pCF}(A,L_1)$ since $I_1\not\models_OA$. Similarly, we have that
$I_2\setminus L_2\not\models_OA$ and $I'_2\models_O\textit{nCF}(A,L_2)$ since $I_2\not\models_OA$.

\begin{proposition}
  Let $P$ be a normal logic program, $L\subseteq\HB_P$ and $M$ a model of the completion of $P$.
  \begin{itemize}
    \item[(1)] $L$ is a loop of $P$ if and only if $L$ is a canonical loop of $\mathcal K=(\emptyset,P)$.
    \item[(2)] $M\models LF(L,P)$ if and only if $M\models_O\cLF(L,I,\mathcal K)$ where $LF(L,P)$ is the loop
        formula associated with $L$ under $P$ \cite{LinZhao:assat}) and $O=\emptyset$.
  \end{itemize}
\end{proposition}
\begin{proof}
  (1) It is obvious since for any atom $h$ there always has an interpretation $I=\{h\}$ such that
      $I\models_O h$ and $I\setminus\{h\}\not\models_O h$.

  (2) $M\models_O\cLF(L,I,\mathcal K)$
    if and only if there is a rule $(r:h\lto\Pos,\Not\Neg)$ in $P$ such that $h\in L$, $\Pos\cap L=\emptyset$,
    $M\models_O\bigwedge_{A\in\Pos}A\wedge\bigwedge_{B\in\Neg}\neg B$ and
    \begin{equation}\label{eq:prop:4.1}
        M\models_O\bigwedge_{A\in\Pos}\delta_1(A,L)\wedge\bigwedge_{B\in\Neg}\neg \delta_2(B,L).
    \end{equation}
    Since $r$ mentions no dl-atoms at all. It  implies that $\delta_1(A,L)=A$ and
    $\delta_2(B,L)=B$. Thus equation (\ref{eq:prop:4.1}) holds
    iff $M\models\bigwedge_{A\in\Pos}A\wedge\bigwedge_{B\in\Neg}\neg B$.
    Consequently, $M\models_O\cLF(L,I,\mathcal K)$ iff
    $M\models LF(L,P)$.
\end{proof}

\begin{proposition}
  Let $\mathcal K=(O,P)$ be a dl-program and $I$ a canonical answer set of $\mathcal K$. Then
  $I$ is minimal in the sense that $\mathcal K$ has no canonical answer set $I'$ such that $I'\subset I$.
\end{proposition}
\begin{proof}
  Suppose there is a canonical answer set $I_1$ of $\mathcal K$ such that $I_1\subset I$. Let $M=I\setminus I_1$.
  Please note that  $I\models_O\comp(\mathcal K)$ and $I_1\models_O\comp(\mathcal K)$. For any atom $h\in M$,
  there is no rule $(h\lto\Pos,\Not\Neg)$ in $P$ such that
  \begin{equation}\label{eq:prop:4.4:a}
    I_1\models_O\bigwedge_{A\in\Pos}A\wedge\bigwedge_{B\in\Neg}\neg B.
  \end{equation}
  Note that there is at least one rule $(h\lto\Pos',\Not\Neg')$ in $P$ such that
  \[I\models_O\bigwedge_{A\in\Pos}A\wedge\bigwedge_{B\in\Neg}\neg B.\]
  It implies that at least one of the following conditions hold:
  \begin{itemize}
    \item There is an atom $h'\in\Pos'$ such that $h'\in M$.
    \item There is a dl-atom $A\in\Pos'$ such that $I_1\not\models_OA$. But note that $I\models_OA$.
    It implies that there is some atom $h'\in I\setminus I_1$, i.e., $h'\in M$, and an interpretation $I^*$
    such that $I^*\not\models_OA$ and $I^*\cup\{h'\}\models_OA$ by (1) of Lemma \ref{lem:cano:dep}.
    \item There is a nonmonotonic dl-atom $B\in\Neg'$ such that $I_1\models_OB$. But note that $I\not\models_OB$.
    It implies that there is some atom $h'\in I\setminus I_1$, i.e., $h'\in M$, and an
    interpretation $I^*$ such that $I^*\models_OB$ and $I^*\cup\{h'\}\not\models_OB$ by (2) of Lemma \ref{lem:cano:dep}.
  \end{itemize}
  It follows that $(h,h')$ is an edge of $G_\mathcal K^c$. Due to that $h$ is an arbitrary atom in $M$ and
  $M$ is finite, there must exists a canonical loop $L$ of $\mathcal K$ such that $L\subseteq M$. We can further assume
  $L$ is such a terminating one, i.e., (a) $L$ is a maximal subset of $M$ and (b) $L$ is a canonical loop of $\mathcal K$ and
  (c) $G_\mathcal K^c$ has no path from one atom of $L$ to an atom of another maximal
  canonical loop $L'$ of $\mathcal K$ with $L'\subseteq M$. Note that $I'\models_O\cLF(L,I,\mathcal K)$ where $I'$
  is the extension of $I$ according to (\ref{IF:formula}). It follows that there is at least one rule
  $(h\lto\Pos'',\Not\Neg'')$ in $P$ such that $h\in L$, $\Pos''\cap L=\emptyset$,
  $$I\models_O\bigwedge_{A\in\Pos''}A\wedge\bigwedge_{B\in\Neg''}\neg B\ \ and\ \
    I'\models_O\bigwedge_{A\in\Pos''}\delta_1(A,L)\wedge\bigwedge_{B\in\Neg''}\neg\delta_2(B,L).$$
  By Lemma \ref{lem:I:I'} and \ref{lem:cDF}, it implies that
  \begin{equation}\label{eq:prop:4.4:b}
   I\setminus L\models_O\bigwedge_{A\in\Pos''}A\wedge\bigwedge_{B\in\Neg''}\neg B.
  \end{equation}

  If $L\subset M$ then $I_1\subset I\setminus L$. In terms of the previous analysis, there is some
  atom $h''\in (I\setminus L)\setminus I_1$, i.e., $h''\in M\setminus L$,
  such that $(h,h'')$ is an edge of $G_\mathcal K^c$. Thus $G_\mathcal K^c$ must have a path
  from $h$ to another canonical loop $L''$ of $\mathcal K$, where $L''\subseteq M$, which contradicts
  with $L$ is a terminating canonical loop. So we have  $L=M$. According
to equation (\ref{eq:prop:4.4:b}),
  we have $I_1\models_O\bigwedge_{A\in\Pos''}A\wedge\bigwedge_{B\in\Neg''}\neg B$
  which contradicts with the condition (\ref{eq:prop:4.4:a}). Consequently, $I_1$ cannot be
  a canonical answer set of $\mathcal K$. Then we complete the proof.
\end{proof}

\begin{proposition}
  Let $\mathcal K=(O,P)$ be a dl-program and $I\subseteq\HB_P$ a canonical answer set of $\mathcal K$.
  Then  $I$ is noncircular.
\end{proposition}
\begin{proof}
  Suppose $I$ is circular, i.e., there exists $M\subseteq I$ such that, for any $(h\lto\Pos,\Not\Neg)$
  in $P$ with $h\in M$ and $I\models_O\bigwedge_{A\in\Pos}A\wedge\bigwedge_{B\in\Neg}\neg B$, the following
  condition holds:
  \begin{equation}\label{eq:can:non:cir:1}
    I\setminus M\not\models_O\bigwedge_{A\in\Pos}A\wedge\bigwedge_{B\in\Neg}\neg B.
  \end{equation}
  Without
  loss of generality, we assume $M$ is such a minimal one. It implies that at least one of the following cases hold:
  \begin{itemize}
    \item $\Pos\cap M\neq\emptyset$ which implies that there is some atom $h'\in\Pos\cap M$.
    \item There is a dl-atom $A\in\Pos$ such that $I\setminus M\not\models_OA$. Knowing that
    $I\models_OA$, it follows that
    there is an interpretation $I^*\subseteq I$ and an atom $h'\in I\setminus (I\setminus M)$,
    i.e., $h'\in M$ such that $I^*\models_OA$ and $I^*\setminus\{h'\}\not\models_OA$ by (1) of Lemma \ref{lem:cano:dep}.
    So that $h'$ is a positive nonmonotonic dependency of $A$.

    \item There is a nonmonotonic dl-atom $B\in\Neg$ such that $I\setminus M\models_OB$. Knowing that
    $I\not\models_OB$, it follows that there is an interpretation $I^*$ and an atom $h'\in I\setminus(I\setminus M)$,
    i.e., $h'\in M$ such that $I^*\models_OB$ and $I^*\cup\{h'\}\not\models_OB$ by (2) of Lemma \ref{lem:cano:dep}.
    So that $h'$ is a negative nonmonotonic dependency of $A$.
  \end{itemize}
  Thus we have that $(h,h')$ is an edge of the canonical dependency graph of $\mathcal K$. Because the atom is an
  arbitrary one in $M$ and $M$ is finite,. there is a terminating canonical loop in
  the generated subgraph of $G_\mathcal K^c$ on $M$, i.e., the graph $G'=(V,E)$ where $V=M$ and $(u,v)\in E$ if
  $(u,v)$ is an edge of $G_\mathcal K^c$. Let $L\subseteq M$ be such a terminating canonical loop.

  Note that $I'\models_O\cLF(L,I,\mathcal K)$ and $L\subseteq I$. It implies that there is
  at least one rule $(h\lto\Pos',\Not\Neg')$ in $P$ such that $h\in L$, $L\cap \Pos'=\emptyset$,
  \[I\models_O\bigwedge_{A\in\Pos'}A\wedge\bigwedge_{B\in\Neg'}\neg B\ \ and\ \
  I'\models_O\bigwedge_{A\in\Pos'}\delta_1(A,L)\wedge\bigwedge_{B\in\Neg'}\neg\delta_2(B,L).\]
  It implies that, by Lemma \ref{lem:I:I'} and Lemma \ref{lem:cDF},
  \[I\setminus L\models_O\bigwedge_{A\in\Pos'}A\wedge\bigwedge_{B\in\Neg'}\neg B.\]

  Thus $I\setminus M\subset I\setminus L$ by equation (\ref{eq:can:non:cir:1}). However,
  using the above analysis, we have that $G'$ has a path from one atom in $L$
  to another loop of $G'$. It contradicts with the fact that $L$ is a terminating canonical loop of $G'$.
  Thus $I$ must be noncircular.
\end{proof}

\begin{proposition}\label{prop:CAS:SAS}
  Let $\mathcal K=(O,P)$ be a dl-program and $I\subseteq\HB_P$ a canonical answer set of $\mathcal K$.
  Then $I$ is a strong answer set of $\mathcal K$.
\end{proposition}
\begin{proof}
  Suppose $I$ is not a strong answer set of $\mathcal K$. Since $I\models\comp(\mathcal K)$, there must
  exist some strong loop $L$ of $\mathcal K$ such that $I'\not\models_O\sLF(L,\mathcal K)$, where
  $I'$ is the extension of $I$ according to (\ref{IF:formula}). It implies that,
  $I'\models_O\bigvee L$ and
  \begin{equation}\label{eq:prop:can:SAS:1}
  I'\not\models_O\bigwedge_{A\in\Pos}\gamma(A,L)\wedge\bigwedge_{B\in\Neg}\neg B
  \end{equation}
  for any rule
  $(h\lto\Pos,\Not\Neg)$ in $P$ with $\Pos\cap L=\emptyset$. Without loss of generality,
  we assume $L$ is a minimal one such that $I'\not\models_O\sLF(L,\mathcal K)$.

  Let $M=L\cap I$. It is evident that $M\neq\emptyset$ and $I\setminus M=I\setminus L$.
  Let $h'$ be an  atom in $M$. Because $h'\in I$, there exists at least one
  rule $(h'\lto\Pos',\Not\Neg')$ in $P$ such that
  \[I\models_O\bigwedge_{A\in\Pos'}A\wedge\bigwedge_{B\in\Neg'}\neg B.\]
  It implies that at least one of the following conditions holds:
  \begin{itemize}
    \item $\Pos'\cap L\neq\emptyset$. It shows that there is some atom $h''\in\Pos'\cap M$.
    \item There is a monotonic dl-atom $A\in\Pos'$ such that $I'\not\models_O\textit{IF}(A,L)$.
    It shows that $I\setminus L\not\models_OA$ by Lemma \ref{lem:I:I'}, i.e., $I\setminus M\not\models_OA$.
    Note that $I\models_OA$. There must have some interpretation $I^*$ and an atom $h''\in I\setminus (I\setminus M)$,
    i.e., $h''\in M$ such that $I^*\models_OA$ and $I^*\setminus\{h''\}\not\models_OA$ by (1) of Lemma \ref{lem:cano:dep}.
  \end{itemize}
  So that $(h',h'')$ is an edge of the canonical dependency graph of $\mathcal K$. Due to the arbitrariness
  of $h'$ and that $M$ is finite, the generated subgraph $G'$ of $G_\mathcal K^c$ on $M$ must
  have a terminating canonical loop $M'$. It is clear that $M'\subseteq M$. Note
  that $M'\subseteq I$ and $I'\models_O\cLF(M',I,\mathcal K)$. It implies that there
  is at least one rule $(h''\lto\Pos'',\Not\Neg'')$ in $P$ such that $h''\in M'$, $\Pos''\cap M'=\emptyset$,
  $$I\models_O\bigwedge_{A\in\Pos''}A\wedge\bigwedge_{B\in\Neg''}\neg B\ \ and\ \
    I'\models_O\bigwedge_{A\in\Pos''}\delta_1(A,M')\wedge\bigwedge_{B\in\Neg''}\neg\delta_2(B,M').$$
  It follows that, by Lemmas \ref{lem:I:I'} and \ref{lem:cDF},
  \[I\setminus M'\models_O\bigwedge_{A\in\Pos''}A\wedge\bigwedge_{B\in\Neg''}\neg B.\]

  However, by equation (\ref{eq:prop:can:SAS:1}) at least one of the following conditions hold:
  \begin{itemize}
    \item $\Pos''\cap L\cap I\neq\emptyset$, i.e., $\Pos''\cap M\neq\emptyset$. It implies that
    there is an atom $h^*\in\Pos''\cap M$ such that $h^*\in M\setminus M'$.
    \item There is a monotonic dl-atom $A\in \Pos''$ such that $I\setminus M\not\models_OA$. But we know
    that $I\setminus M'\models_OA$ by Lemma \ref{lem:cDF}. It shows that there is an atom $h^*\in M\setminus M'$
    such that $I^*\not\models_OA$ and $I^*\cup\{h^*\}\models_OA$ for some interpretation $I^*$ by Lemma \ref{lem:cano:dep}.
  \end{itemize}
  It follows that $(h'',h^*)$ is also an edge of $G_\mathcal K^c$. Since $M$ is finite, $G'$ must have
  a path from $h''$ to another canonical loop of $G'$. It contradicts with $M'$ is a terminating
  canonical loop of $G'$. Consequently, $I$ is a strong answer set of $\mathcal K$.
\end{proof}

\begin{proposition}
  Let $\mathcal K=(O,P)$ be a dl-program in which $P$ does not mention the operator $\ominus$.
  Then $I\subseteq\HB_P$ is a canonical answer set of $\mathcal K$ if and only if $I$ is
  a strong answer set of $\mathcal K$.
\end{proposition}
\begin{proof}
  By Proposition \ref{prop:CAS:SAS}, it is sufficient to show that if $I$ is a strong
  answer set of $\mathcal K$ then $I$ is a canonical answer set of $\mathcal K$. Suppose
  $I$ is a strong answer set of $\mathcal K$ but $I$ is not a canonical answer set of $\mathcal K$.
  Since $I\models\comp(\mathcal K)$, it implies that there exists at least one canonical loop
  $L$ of $\mathcal K$ such that $I'\not\models_O\cLF(L,I,\mathcal K)$, where $I'$ is the extension of $I$
  according to (\ref{IF:formula}).

  Since $P$ mentions no $\ominus$, all dl-atoms appearing in $P$
  must be monotonic. In particular, if $A$ is a monotonic dl-atom and there is some atom $p(\vec c)$ and
  an interpretation $I^*$ such that $I^*\not\models_OA$ and $I^*\cup\{p(\vec c)\}\models_OA$ then
  $A$ must contain $S\oplus p$ (or $S\odot p$) for some $S$. It implies that $L$ is also
  a strong loop of $\mathcal K$ and then $I'\models_O\sLF(L,\mathcal K)$, i.e.,
  $P$ has at least one rule $(h\lto\Pos,\Not\Neg)$ such that $h\in L$, $\Pos\cap L=\emptyset$ and
  \[I'\models_O\bigwedge_{A\in\Pos}\gamma(A,L)\wedge\bigwedge_{B\in\Neg}\neg B.\]
  Note that no dl-atoms mention $\ominus$. By Lemma \ref{lem:I:I'}, it follows that
  \[I'\models_O\bigwedge_{A\in\Pos}\delta_1(A,L)\wedge\bigwedge_{B\in\Neg}\neg\delta_2(B,L)\ \ and\ \
    I\models_O \bigwedge_{A\in\Pos}A\wedge\bigwedge_{B\in\Neg}\neg B.\]
  It contradicts with $I'\not\models_O\cLF(L,I,\mathcal K)$. Thus $I$ is a
  canonical answer set of $\mathcal K$.
\end{proof}

\end{document}